\documentclass[11pt]{article}

\usepackage[in]{fullpage}
\usepackage{times}
\usepackage{graphicx}
\usepackage[rflt]{floatflt}
\usepackage{epsfig,subfigure,color,multirow}
\usepackage{amsmath,amssymb,algorithm,algorithmic,theorem,float,bbm,bm,enumerate}
\usepackage{url}
\usepackage{epstopdf}

\newcommand{\beq}{\begin{equation}}
\newcommand{\eeq}{\end{equation}}

\newcommand\R{\mathbb{R}}

\renewcommand{\a}{\mathbf{a}}
\renewcommand{\b}{\mathbf{b}}
\renewcommand{\c}{\mathbf{c}}

\newcommand{\q}{\mathbf{q}}

\renewcommand{\u}{\mathbf{u}}
\renewcommand{\v}{\mathbf{v}}
\newcommand{\w}{\mathbf{w}}
\newcommand{\x}{\mathbf{x}}
\newcommand{\y}{\mathbf{y}}
\newcommand{\z}{\mathbf{z}}

\newcommand{\cX}{{\cal X}}

\newcommand{\cZ}{{\cal Z}}

\newcommand{\bM}{\mathbf{M}}

\newcommand{\bA}{\mathbf{A}}
\newcommand{\bB}{\mathbf{B}}

\newcommand{\bI}{\mathbf{I}}
\newcommand{\bD}{\mathbf{D}}
\newcommand{\bQ}{\mathbf{Q}}

\newcommand{\E}{\mathbf{E}}

\newcommand{\myref}[1]{(\ref{#1})}

\DeclareMathOperator{\argmin}{argmin}

\DeclareMathOperator{\ri}{ri}

\newcounter{exampleI}
{\theorembodyfont{\rmfamily} \theoremstyle{plain} }

\newcounter{exampleII}
{\theorembodyfont{\rmfamily} \theoremstyle{plain} }

\newcounter{exampleIII}
{\theorembodyfont{\rmfamily} \theoremstyle{plain} }

{\theorembodyfont{\rmfamily} }
{\theorembodyfont{\rmfamily} }
{\theorembodyfont{\rmfamily} \newtheorem{rmk}{Remark}}
\newtheorem{thm}{Theorem}

\newtheorem{lem}{Lemma}
\newtheorem{cor}{Corollary}

\newtheorem{asm}{Assumption}
\newcommand{\proof}{\noindent{\itshape Proof:}\hspace*{1em}}

\newcommand{\qed}{\nolinebreak[1]~~~\hspace*{\fill} \rule{5pt}{5pt}\vspace*{\parskip}\vspace*{1ex}}

\newcommand {\commentout}[1] {}

\title{Online Alternating Direction Method}
\author{Huahua Wang \\
Dept of Computer Science \& Engg\\
University of Minnesota, Twin Cities\\
huwang@cs.umn.edu
\and
Arindam Banerjee\\
Dept of Computer Science \& Engg\\
University of Minnesota, Twin Cities\\
banerjee@cs.umn.edu
}

\date{}

\begin{document}

\maketitle

\begin{abstract}
Online optimization has emerged as powerful tool in large scale optimization. In this paper, we introduce efficient online optimization algorithms based on the alternating direction method (ADM), which can solve online convex optimization under linear constraints where the objective could be non-smooth. We introduce new proof techniques for ADM in the batch setting, which yields a $O(1/T)$ convergence rate for ADM and forms the basis for regret analysis in the online setting. We consider two scenarios in the online setting, based on whether an additional Bregman divergence is needed or not. In both settings, we establish regret bounds for both the objective function as well as constraints violation for general and strongly convex functions. We also consider inexact ADM updates where certain terms are linearized to yield efficient updates and show the stochastic convergence rates. In addition, we briefly discuss that online ADM can be used as projection-free online learning algorithm in some scenarios. Preliminary results are presented to illustrate the performance of the proposed algorithms.
\end{abstract}

\section{Introduction}

In recent years, online optimization~\cite{celu06,Zinkevich03,haak07} and its batch counterpart stochastic gradient descent~\cite{Robi51:SP,Judi09:SP} has contributed substantially to advances in large scale optimization techniques for machine learning. Online convex optimization has been generalized to handle time-varying and non-smooth convex functions~\cite{Duchi10_comid,duchi09,xiao10}. Distributed optimization, where the problem is divided into parts on which progress can be made in parallel, has also contributed to advances in large scale optimization~\cite{boyd10,Bertsekas89,ceze98}.

Important advances have been made based on the above ideas in the recent literature. Composite objective mirror descent (COMID)~\cite{Duchi10_comid} generalizes  mirror descent~\cite{Beck03} to the online setting. COMID also includes certain other proximal splitting methods such as FOBOS~\cite{duchi09} as special cases. Regularized dual averaging (RDA)~\cite{xiao10}  generalizes dual averaging~\cite{nesterov09} to online and composite optimization, and can be used for distributed optimization~\cite{Duchi11_dv}. The three methods consider the following composite objective optimization~\cite{nest07:composite}:
\begin{align} \label{eq:co_pm}
\min_{\x\in \mathcal{X}} \sum_{t=1}^T (f_t(\x) + g(\x))~,
\end{align}
where the functions $f_t,g$ are convex functions and $\mathcal{X}$ is a convex set. Solving \myref{eq:co_pm} usually involves the projection onto $\mathcal{X}$. In some cases, e.g.,  when $g$ is the $\ell_1$ norm or $\mathcal{\cX}$ is the unit simplex, the projection can be done efficiently.  In general, the full projection requires an inner loop algorithm, leading to a double loop algorithm for solving~\myref{eq:co_pm}~\cite{hazan12:free}.

In this paper, we propose single loop online optimization algorithms for composite objective optimization subject to linear constraints. In particular, we consider optimization problems of the following form:
\beq \label{eq:admm_pm}
\min_{\x\in \mathcal{X},\z\in\mathcal{Z}} \sum_{t=1}^T (f_t(\x) + g(\z)) \quad \text{s.t.} \quad \bA\x + \bB \z = \c~,
\eeq
where $\bA \in \R^{m\times n_1}, \bB \in \R^{m\times n_2}, \c \in \R^m$, $\x \in \mathcal{X} \in \R^{n_1\times 1}, \z \in \mathcal{Z} \in \R^{n_2\times 1}$ and $\mathcal{X}$ and $\mathcal{Z}$ are convex sets. The linear equality constraint introduces splitting variables and thus splits functions and feasible sets into simpler constraint sets $\x \in \mathcal{X}$ and $\z \in \mathcal{Z}$. \myref{eq:admm_pm} can easily accommodate linear inequality constraints
by introducing a slack variable, which will be discussed in Section~\ref{sec:linie}. In the sequel, we drop the convex sets $\mathcal{X}$ and $\mathcal{Z}$ for ease of exposition, noting that one can consider $g$ and other additive functions to be the indicators of suitable convex feasible sets. $f_t$ and $g$ can be non-smooth, including piecewise linear and indicator functions. In the context of machine learning, $f_t$ is usually a loss function such as $\ell_1, \ell_2$, hinge and logistic loss, while $g$ is a regularizer, e.g., $\ell_1$, $\ell_2$, nuclear norm, mixed-norm and total variation.


In the batch setting, where $f_t = f$, \myref{eq:admm_pm} can be solved by the well known alternating direction method of multipliers (ADMM or ADM)~\cite{boyd10}.
First introduced in~\cite{Gabay76}, ADM has since been extensively explored in recent years due to its ease of applicability and empirical performance in a wide variety of problems, including composite objectives~\cite{boyd10,Eckstein92,Lin_Ma09}.  It has been shown as a special case of Douglas-Rachford splitting method \cite{comb09:prox,Douglas56, Eckstein92}, which in turn is a special case of the  proximal point method \cite{Rockafellar76}. Recent literature has illustrated the empirical efficiency of ADM in a broad spectrum of applications ranging from image processing \cite{Ng10,Figueiredo10,Afonso10:tv,Chan11} to applied statistics and machine learning~\cite{Scheinberg10,Afonso10:tv,Yuan09,Yuan09b,yang09, Lin_Ma09,Barman11,Meshi10, Martins11}. ADM has been shown to outperform state-of-the-art methods for sparse problems, including LASSO~\cite{Tibshirani96,Hastie09,Afonso10:tv,boyd10}, total variation~\cite{Goldstein10:tv}, sparse inverse covariance selection~\cite{Dempster72,Banerjee08,Friedman08,Meinshausen06,Scheinberg10,Yuan09}, and sparse and low rank approximations \cite{Yuan09b,Lin_Ma09,cand09:rpca}. ADM have also been used to solve linear programs (LPs)~\cite{Eckstein90}, LP decoding~\cite{Barman11} and MAP inference problems in graphical models~\cite{Martins11,Meshi10,wang12:kladm}. In addition, an advantage of ADM is that it can handle linear equality constraint of the form $\{ \x,\z| \bA \x + \bB \z = \c\}$, which makes distributed optimization by variable splitting in a batch setting straightforward~\cite{Bertsekas89,Nedic10,boyd10,boyd12:gpbs,Giannakis07}.
For further understanding of ADM, we refer the readers to the comprehensive review by~\cite{boyd10} and references therein.

Although the proof of global convergence of ADM can be found in \cite{Gabay83,Eckstein92,boyd10}, the literature does not have the convergence rate for ADM \footnote{During/after the publication of our preliminary version~\cite{wang12:oadm}, the convergence rate for ADM  was shown in ~\cite{he12:vi,he12:cst,luo12:admm,deng12:admm,dan12:admm,Goldstein12:fadmm}, but our proof is different and self-contained. In particular, the other approaches do not prove the convergence rate for the objective, which is the key for the regret analysis in the online setting or stochastic setting.} or even the convergence rate for the objective, which is fundamentally important to regret analysis in the online setting. We introduce new proof techniques for the rate of convergence of ADM in the batch setting, which establish a $O(1/T)$ convergence rate for the objective, the optimality conditions (constraints) and ADM based on variational inequalities~\cite{fapa03}. The $O(1/T)$ convergence rate for ADM is in line with gradient methods for composite objective~\cite{Nest04:bkcov,nest07:composite,duchi09}\footnote{ The gradient methods can be accelerated to achieve the $O(1/T^2)$ convergence rate~\cite{Nest04:bkcov,nest07:composite}.}. Our proof requires rather weak assumptions compared to the Lipschitz continuous gradient required in general in gradient methods~\cite{Nest04:bkcov,nest07:composite,duchi09}. Further, the convergence analysis for the batch setting forms the basis of regret analysis in the online setting.




In an online or stochastic gradient descent setting, where $f_t$ is a time-varying function, \myref{eq:admm_pm} amounts to solving a sequence of equality-constrained subproblems, which in general leads to a double-loop algorithm where the inner loop ADM iterations have to be run till convergence after every new data point or function is revealed. As a result, ADM has not yet been generalized to the online setting.

We consider two scenarios in the online setting, based on whether an additional Bregman divergence is needed or not for a proximal function in each step. We propose efficient online ADM (OADM) algorithms for both scenarios which make a single pass through the update equations and avoid a double loop algorithm. In the online setting, while a single pass through the ADM update equations is not guaranteed to satisfy the linear constraint $\bA\x + \bB \z = \c$ in each iteration, we consider two types of regret: regret in the {\em objective} as well as regret in {\em constraint violation}. We establish both types of regret bounds for general and strongly convex functions. In Table \ref{main}, we summarize the main results of OADM and also compare with OGD~\cite{Zinkevich03}, FOBOS~\cite{duchi09}, COMID~\cite{Duchi10_comid} and RDA~\cite{xiao10}. While OADM aims to solve linearly-constrained composite objective optimization problems, OGD, FOBOS and RDA are for such problems without explicit constraints. In both general and strongly convex cases, our methods achieve the optimal regret bounds for the objective as well as the constraint violation, while start-of-the-art methods achieve the optimal regret bounds for the objective. We also present preliminary experimental results illustrating the performance of the proposed OADM algorithms in comparison with FOBOS and RDA~\cite{duchi09,xiao10}.

The key advantage of the OADM algorithms can be summarized as follows: Like COMID and RDA, OADM can solve online composite optimization problems, matching the regret bounds for existing methods. The ability to additionally handle linear equality constraint of the form $\bA \x + \bB \z = \c$ makes non-trivial variable splitting possible yielding efficient distributed online optimization algorithms \cite{Dekel12_minbatch} and projection-free online learning \cite{hazan12:free} based on OADM.  Further, the notion of regret in both the objective as well as constraint may contribute towards development of suitable analysis tools for online constrained optimization problems~\cite{Mannor06,Majin11}.

\begin{table}\label{main}
\centering
\begin{tabular}{|c||c|c||c|}
\hline
Problem  & \multicolumn{2}{c||}{ $\underset{\bA\x + \bB\z = \c}{\min} \sum_t f_t(\x) + g(\z)$ } & $\min_{\x} \sum_t f_t(\x) + g(\x)$\\
\hline
Methods  &  \multicolumn{2}{c||}{ OADM } & OGD, FOBOS, COMID, RDA \\
\hline
Regret Bounds & Objective & constraint& Objective \\
\hline
General Convex &  $O(\sqrt{T})$ & $O(\sqrt{T})$ & $O(\sqrt{T})$ \\
\hline
Strongly Convex & $ O(\log{(T)})$ & $O(\log{(T)})$ & $ O(\log{(T)}) $ \\
\hline
\end{tabular}
\caption{Main results for regret bounds of OADM in solving linearly-constrained composite objective optimization, in comparison with OGD, FOBOS, COMID and RDA in solving composite objective optimization. In both general and strongly convex cases, OADM achieves the optimal regret bounds for the objective, matching the results of the state-of-the-art methods. In addition, OADM also achieves the optimal regret bounds for constraint violation, showing the equality constraint will be satisfied on average in the long.}
\end{table}

We summarize our main contributions as follows:
\begin{itemize}
\item We establish a $O(1/T)$ convergence rate for the objective, optimality conditions (constraints) and variational inequality for ADM.
\item We propose online ADM (OADM), which is the first single loop online algorithm to explicitly solve the linearly-constrained problem \myref{eq:admm_pm} by just doing a single pass over examples.  
\item In OADM, we establish the optimal regret bounds for both objective and constraint violation for general as well as strongly convex functions. The introduction of regret for constraint violation which allows constraints to be violated at each round but guarantees constraints to be satisfied on average in the long run. 
\item We show some inexact updates in the OADM through the use of an additional Bregman divergence, including OGD and COMID as special cases. For OADM with inexact updates, we also show the stochastic convergence rates. 
\item For an intersection of simple constraints, e.g., linear constraint (simplex), OADM is a projection-free online learning algorithm achieving the optimal regret bounds for both general and strongly convex functions. 
\end{itemize}

The rest of the paper is organized as follows. In Section 2, we analyze batch ADM and establish its convergence rate. In Section 3, we propose OADM to solve the online optimization problem with linear constraints. In Sections 4 and 5, we present the regret analysis in two different scenarios based on whether an additional Bregman divergence is added or not. In Section 6, we discuss inexact ADM updates and show the stochastic convergence rates, show the connection to related works and  projection-free online learning based on OADM. We present preliminary experimental results in Section 7, and conclude in Section 8.


\section{Analysis for Batch Alternating Direction Method}\label{sec:admm}
We consider the batch ADM problem where $f_t$ is fixed in \myref{eq:admm_pm}, i.e.,
\begin{align} \label{eq:badm_pm}
\min_{\x,\z} f(\x) + g(\z) \quad \text{s.t.} \quad \bA\x + \bB \z = \c~.
\end{align}
The Lagrangian~\cite{boyd04:convex,Bertsekas99} for the equality-constrained optimization problem \myref{eq:badm_pm} is
\begin{equation}\label{eq:lag1}
L(\x,\y,\z) = f(\x) + g(\z) + \langle \y, \bA\x + \bB\z - \c \rangle,
\end{equation}
where $\x,\z$ are the primal variables and $\y$ is the dual variable. To penalize the violation of equality constraint, augmented Lagrangian methods use an additional quadratic penalty term.
In particular, the augmented Lagrangian~\cite{Bertsekas99} for \myref{eq:admm_pm} is
\begin{equation}\label{eq:alag}
L_{\rho}(\x,\y,\z) = f(\x) + g(\z) + \langle \y, \bA\x + \bB\z - \c \rangle + \frac{\rho}{2} \| \bA\x + \bB\z - \c \|^2,
\end{equation}
where $\rho > 0 $ is a penalty parameter.
Batch ADM updates the three variables $(\x,\z,\y)$ by alternatingly minimizing the augmented Lagrangian.
It executes the following three steps iteratively till convergence~\cite{boyd10}:
\begin{align}
&\x_{t+1} = \underset{\x}{\argmin} \left \{ f(\x) + \langle \y_t, \bA\x + \bB\z_t - \c \rangle + \frac{\rho}{2} \| \bA\x + \bB\z_t -\c \|^2 \right\}~,  \label{eq:admm_x}\\
&\z_{t+1}  = \underset{\z}{\argmin} \left \{ g(\z) + \langle \y_t, \bA\x_{t+1} + \bB\z - \c \rangle + \frac{\rho}{2} \| \bA\x_{t+1} + \bB\z - \c \|^2 \right \}~,  \label{eq:admm_z} \\
&\y_{t+1}  = \y_t + \rho(\bA\x_{t+1} + \bB\z_{t+1} - \c)~. \label{eq:admm_y}
\end{align}
At step $(t+1)$, the equality constraint in~\myref{eq:badm_pm} is not necessarily satisfied in ADM. However, one can show that the equality constraint is satisfied in the long run so that $\lim_{t\rightarrow \infty} \bA\x_t + \bB\z_t - \c \rightarrow \mathbf{0}$.

While global convergence of ADMM has been established under appropriate conditions, we are interested in the rate of convergence of ADM in terms of iteration complexity, i.e., the number of iterations needed to obtain an $\epsilon$-optimal solution. Most first-order methods require functions to be smooth or having Lipschitz continuous gradient to establish the convergence rate~\cite{Nest04:bkcov,nest07:composite,duchi09}. The assumptions in establishing convergence rate of ADM are relatively simple~\cite{boyd10}, and are stated below for the sake of completeness:
\begin{asm}\label{asm:adm}
\quad

(a) $f:\R^{n_1}\rightarrow\R\cup\{+\infty\}$ and $g:\R^{n_2}\rightarrow\R\cup\{+\infty\}$ are closed, proper and convex.

(b) An optimal solution to \myref{eq:badm_pm} exists. Let $\{\x^*,\z^*,\y^*\}$ be an optimal solution. Denote $\| \y^* \|_2 = D_{\y}, \| \z^* \|_2 = D_{\z}$.

(c) Without loss of generality, $\z_0 = 0, \y_0 = 0$. Let $\lambda_{\max}^{\bB}$ be the largest eigenvalue of $\bB^T\bB$.
\end{asm}


We first analyze the convergence rate for the objective and optimality conditions (constraints) separately using new proof techniques, which play an important role for the regret analysis in the online setting. Then, a joint analysis of the objective and constraints using a variational inequality~\cite{fapa03} establishes the $O(1/T)$ convergence rate for ADM. 


%

\subsection{Convergence Rate for the Objective}
The updates of $\x,\z$ implicitly generate the (sub)gradients of $f(\x_{t+1})$ and $g(\z_{t+1})$, as given in the following lemma.
\begin{lem} \label{lem:sg}
Let $\partial f(\x_{t+1})$ be the subgradient of $f(\x)$ at $\x_{t+1}$, we have
\begin{align}
-\bA^T(\y_t + \rho(\bA\x_{t+1} + \bB\z_t - \c)) &\in \partial f(\x_{t+1})\label{eq:fg1}~, \\
-\bA^T(\y_{t+1} + \rho(\bB\z_t - \bB\z_{t+1})) &\in \partial f(\x_{t+1})\label{eq:fg2}
\end{align}
Let $\partial g(\z_{t+1})$ be the subgradient of $g(\z)$ at $\z_{t+1}$, we have
\begin{align}
&-\bB^T \y_{t+1}  \in \partial g(\z_{t+1})~. \label{eq:gy}
\end{align}
\end{lem}
\proof
Since $\x_{t+1}$ minimizes \myref{eq:admm_x}, we have
\begin{align*}
0 \in \partial f(\x_{t+1}) + \bA^T\y_t + \rho\bA^T(\bA\x_{t+1} - \bB\z_t - \c)~.
\end{align*}
Rearranging the terms gives \myref{eq:fg1}. using \myref{eq:admm_y} yield \myref{eq:fg2}.

Similarly, $\z_{t+1}$ minimizes \myref{eq:admm_z}, then
\begin{align*}
\partial g(\z_{t+1}) + \bB^T\y_t + \rho\bB^T(\bA\x_{t+1} + \bB\z_{t+1} - \c) \in 0~.
\end{align*}
Rearranging the terms and using \myref{eq:admm_y} yield \myref{eq:gy}.
\qed

The following lemma shows the inaccuracy of the objective with respect to the optimum at $(t+1)$ is bounded by step differences of $\y$ and $\z$.
\begin{lem}\label{lem:bd1}
Let the sequences $\{\x_t, \z_t,\y_t\}$ be generated by ADM. Then for any $\x^*,\z^*$ satisfying $\bA\x^* + \bB\z^* = \c$, we have
\begin{align}\label{eq:lem_bd}
& f(\x_{t+1}) + g(\z_{t+1}) - (f(\x^*) + g(\z^*)) \nonumber \\
&\leq \frac{1}{2\rho}(\| \y_{t} \|_2^2 - \| \y_{t+1} \|_2^2)
- \frac{\rho}{2} \| \bA\x_{t+1} + \bB\z_t - \c\|_2^2 + \frac{\rho}{2}  ( \|\bB\z^* - \bB\z_t\|_2^2 -  \|\bB\z^* - \bB\z_{t+1}\|_2^2)~.
\end{align}
\end{lem}
\proof
Since $f(\x)$ is a convex function and its subgradient is given in \myref{eq:fg2},
\begin{align} \label{eq:cf}
 f(\x_{t+1}) - f(\x^*) & \leq -  \langle \bA^T(\y_{t+1} + \rho (\bB\z_t - \bB\z_{t+1})), \x_{t+1} - \x^*  \rangle \nonumber \\
& = -  \langle \y_{t+1} + \rho (\bB\z_t - \bB\z_{t+1}), \bA\x_{t+1} - \bA\x^*  \rangle \nonumber \\
& = -  \langle \y_{t+1}, \bA\x_{t+1} - \c + \bB\z^* \rangle + \rho \langle \bB\z_{t+1} - \bB\z_t, \bA\x_{t+1} - \c + \bB\z^*  \rangle \nonumber \\
& = -  \langle \y_{t+1}, \bA\x_{t+1} - \c + \bB\z^* \rangle + \frac{\rho}{2} (\|\bB\z^* - \bB\z_t\|_2^2 - \|\bB\z^* - \bB\z_{t+1}\|_2^2 \nonumber \\
&\quad + \|\bA\x_{t+1} + \bB\z_{t+1}-\c\|_2^2 - \|\bA\x_{t+1} + \bB\z_{t}-\c\|_2^2 ) ~.
\end{align}
where the last equality uses 
\begin{align}\label{eq:four}
\langle \u_1 - \u_2, \u_3 + \u_4 \rangle = \frac{1}{2}(\| \u_4 - \u_2 \|_2^2 - \|\u_4 - \u_1\|_2^2 + \| \u_3 +\u_1 \|_2^2 - \| \u_3 + \u_2 \|_2^2).
\end{align}
Similarly, for convex function $g(\z)$, using its subgradient in \myref{eq:gy}, we have
\begin{align} \label{eq:cg}
&g(\z_{t+1}) - g(\z^*) \leq -  \langle \bB^T\y_{t+1}, \z_{t+1} - \z^*  \rangle = -  \langle \y_{t+1}, \bB\z_{t+1} - \bB\z^*  \rangle~.
\end{align}
Adding \myref{eq:cf} and \myref{eq:cg} together yields
\begin{align}\label{eq:lem_bd1}
&f(\x_{t+1}) + g(\z_{t+1}) - (f(\x^*) + g(\z^*))  \nonumber \\
& \leq - \langle \y_{t+1}, \bA\x_{t+1} + \bB\z_{t+1} - \c \rangle  + \frac{\rho}{2} \| \bA\x_{t+1} + \bB\z_{t+1} - \c\|_2^2   \nonumber \\
&  - \frac{\rho}{2} \| \bA\x_{t+1} + \bB\z_t - \c\|_2^2
+ \frac{\rho}{2}  ( \|\bB\z^* - \bB\z_t\|_2^2 -  \|\bB\z^* - \bB\z_{t+1}\|_2^2)~.
\end{align}
Recalling \myref{eq:admm_y}, the first two terms in \myref{eq:lem_bd1} can be rewritten as
\begin{align}\label{eq:yycancel}
& - \langle \y_{t+1}, \bA\x_{t+1} + \bB\z_{t+1} - \c \rangle + \frac{\rho}{2} \| \bA\x_{t+1} + \bB\z_{t+1} - \c \|_2^2 \nonumber \\
& = \frac{1}{2\rho} (2\langle \y_{t+1}, \y_t - \y_{t+1}\rangle + \| \y_t - \y_{t+1}\|_2^2)  \nonumber \\
& = \frac{1}{2\rho}(\| \y_{t} \|_2^2 - \| \y_{t+1} \|_2^2)~.
\end{align}
Plugging back into \myref{eq:lem_bd1} yields the result.
\qed

As observed in several experiments~\cite{boyd10}, the objective is not monotonically non-increasing. The following theorem shows the objective of ADM has the $O(1/T)$ convergence rate in an ergodic sense.
\begin{thm} \label{thm:bobj}
Let the sequences $\{\x_t, \z_t,\y_t\}$ be generated by ADM and $\bar{\x}_T = \frac{1}{T}\sum_{t=1}^T \x_t, \bar{\z}_T = \frac{1}{T}\sum_{t=1}^T \z_t$. For any $\x^*,\z^*$ satisfying $\bA\x^* + \bB\z^* = \c$, for any $T$, we have
\begin{align}
&f(\bar{\x}_{T}) + g(\bar{\z}_{T}) - (f(\x^*) + g(\z^*))
 \leq \frac{\lambda_{\max}^{\bB}D_{\z}^2\rho}{2T}~. \label{eq:thm_bobj}
\end{align}
\end{thm}
\proof
In \myref{eq:lem_bd}, ignoring $- \frac{\rho}{2} \| \bA\x_{t+1} + \bB\z_t - \c\|_2^2$ and summing over $t$ from $0$ to $T-1$, we have the following telescoping sum
\begin{align*}
&\sum_{t=0}^{T-1}\left[f(\x_{t+1}) + g(\z_{t+1}) - (f(\x^*) + g(\z^*))\right] \nonumber \\
&\leq \frac{1}{2\rho}(\| \y_{0} \|_2^2 - \| \y_{T} \|_2^2) + \frac{\rho}{2}  ( \|\bB\z^* - \bB\z_0\|_2^2 -  \|\bB\z^* - \bB\z_{T}\|_2^2)~.
\end{align*}
Since both $f$ and $g$ are convex, dividing by $T$, applying Jensen's inequality and letting the assumptions hold complete the proof.
\qed

Although \myref{eq:thm_bobj} shows that the objective value converges to the optimal value, $\{\x_{t+1},\z_{t+1}\}$ may not be feasible and the equality constraint may not necessarily be satisfied. 
\subsection{Convergence Rate for the Optimality Conditions (Constraints)}
Assume that $\{ \x^*, \z^*,\y^*\}$ satisfies the KKT conditions of the Lagrangian \myref{eq:lag1}, i.e.,
\begin{align}
-\bA^T\y^* & \in \partial f(\x^*)~, \label{eq:kkdx}\\
-\bB^T\y^* & \in \partial g(\z^*)~,\label{eq:kkdz}\\
\bA\x^* + \bB\z^* - \c & = 0~. \label{eq:kkdy}
\end{align}

According to~\myref{eq:fg2}, conditio~\myref{eq:kkdx} holds if $\bB\z_{t+1} - \bB\z_t = 0$. According to \myref{eq:gy}, condition \myref{eq:kkdz} holds for every iterate. Therefore, the KKT conditions~\myref{eq:kkdx}-\myref{eq:kkdy} hold if the following optimality conditions are satisfied:
\begin{align}
\bB\z_{t+1} - \bB\z_{t} &= 0~, \label{eq:pr} \\
\bA\x_{t+1} + \bB\z_{t+1} - \c &= 0 \label{eq:ec}~,
\end{align}
The LHS of \myref{eq:pr} is called \emph{primal residual} and the LHS of~\myref{eq:ec} is called  equality constraint violation or \emph{dual residual}~\cite{boyd10} when considering~\myref{eq:admm_y}. 

Define a residual function of optimality conditions as
\begin{align}
R(s,t) = \| \bA\x_{s} + \bB\z_{t} - \c \|_2^2 + \| \bB\z_{t} - \bB\z_{s-1} \|_2^2~,
\end{align}
where $s \in \{t,t+1\}$. In particular, the residual after the $\z$ update~\myref{eq:admm_z} at iteration $(t+1)$ is
\begin{align}
R(t+1,t+1) = \| \bA\x_{t+1} + \bB\z_{t+1} - \c \|_2^2 + \| \bB\z_{t+1} - \bB\z_t \|_2^2~.
\end{align}
and the residual after the $\x$-update~\myref{eq:admm_x} at $(t+1)$ is
\begin{align}
R(t+1,t) =\| \bA\x_{t+1} + \bB\z_t - \c \|_2^2~.
\end{align}
Therefore, the convergence of $R(t+1,t+1)$ implies the convergence of the optimality conditions.

The following two lemmas show the residuals of optimality conditions (constraints) are monotonically non-increasing.
\begin{lem}\label{lem:flbd}
Let the sequences $\{\x_t, \z_t,\y_t\}$ be generated by ADM. Then
\begin{align}
R(t+1,t) \leq R(t,t)
\end{align}
\end{lem}
\proof
Since $f(\x)$ is a convex function and its subgradient is given in \myref{eq:fg1}, for any $\x$, we have
\begin{align}\label{eq:t1f}
f(\x_{t+1}) - f(\x) & \leq - \langle \bA^T(\y_{t} + \rho(\bA\x_{t+1} + \bB\z_t - \c)), \x_{t+1} - \x \rangle \nonumber \\
& = \langle \y_{t}, \bA\x - \bA\x_{t+1} \rangle + \rho \langle \bA\x_{t+1} + \bB\z_t - \c , \bA\x - \bA\x_{t+1} \rangle~.
\end{align}
Letting $\x = \x_t$, we have
\begin{align}\label{eq:t1ft}
&f(\x_{t+1}) - f(\x_t) \leq \langle \y_{t}, \bA\x_t - \bA\x_{t+1} \rangle + \rho \langle \bA\x_{t+1} + \bB\z_t - \c , \bA\x_t - \bA\x_{t+1} \rangle \nonumber \\
& = \langle \y_{t} , \bA\x_t - \bA\x_{t+1} \rangle + \frac{\rho}{2}( \| \bA\x_t + \bB\z_t - \c \|_2^2 - \|\bA\x_{t+1} + \bB\z_t - \c \|_2^2 - \|\bA\x_t - \bA\x_{t+1} \|_2^2 )~.
\end{align}
where the last equality uses
\begin{align}\label{eq:three}
\langle \u_1 - \u_2, \u_3 - \u_1 \rangle = \frac{1}{2}( \| \u_2 - \u_3 \|_2^2 - \| \u_1 - \u_2\|_2^2 - \|\u_1 - \u_3\|_2^2 )~.
\end{align}
Using the subgradient of $f$ given in~\myref{eq:fg2} at $\x_t$, for any $\x$,
\begin{align}
f(\x_{t}) - f(\x) \leq - \langle \bA^T(\y_{t} + \rho(\bB\z_{t-1} - \bB\z_{t})), \x_{t} - \x \rangle ~.
\end{align}
Letting $\x = \x_{t+1}$, we have
\begin{align}\label{eq:tft1}
f(\x_{t}) - f(\x_{t+1}) & \leq -\langle \y_{t}, \bA\x_t - \bA\x_{t+1} \rangle + \rho \langle \bB\z_{t-1} - \bB\z_{t}, \bA\x_{t+1} - \bA\x_t \rangle \nonumber \\
& \leq \langle \bA\x_{t+1} - \bA\x_t, \y_{t} \rangle + \frac{\rho}{2}( \|\bA\x_{t+1} - \bA\x_t\|_2^2 + \| \bB\z_{t-1} - \bB\z_{t} \|_2^2 )~.
\end{align}
Adding \myref{eq:t1ft} and \myref{eq:tft1} together and rearranging the terms complete the proof.
\qed

\begin{lem}\label{lem:pdr}
Let the sequences $\{\x_t, \z_t,\y_t\}$ be generated by ADM. Then
\begin{align}\label{eq:cst}
R(t+1,t+1) \leq R(t+1,t)
\end{align}
\end{lem}
\proof
Recalling the subgradient of convex function $g(\z)$ given in \myref{eq:gy}, we have
\begin{align}
&g(\z_{t+1}) - g(\z_t) \leq  \langle - \bB^T\y_{t+1}, \z_{t+1} - \z_{t} \rangle \label{eq:g1}~, \\
&g(\z_t) - g(\z_{t+1}) \leq \langle - \bB^T\y_{t}, \z_t - \z_{t+1} \rangle \label{eq:g2}~.
\end{align}
Adding \myref{eq:g1} and \myref{eq:g2} together yields
\begin{align} \label{eq:pos}
0 \leq \langle \bB^T(\y_{t+1} - \y_t), \z_t - \z_{t+1} \rangle = \langle \y_{t+1} - \y_t, \bB\z_t - \bB\z_{t+1} \rangle~.
\end{align}
According to \myref{eq:admm_y}, the right-hand side can be rewritten as
\begin{align}
& \langle \y_{t+1} - \y_t, \bB\z_t - \bB\z_{t+1} \rangle \nonumber \\
& = \rho  \langle \bA\x_{t+1} + \bB\z_{t+1} - \c, (\bB\z_t -\c) - (\bB\z_{t+1} - \c) \rangle \nonumber \\
& = \frac{\rho }{2}\big( \| \bA\x_{t+1} + \bB\z_t - \c\|_2^2 - \| \bB\z_{t+1} - \bB\z_t\|_2^2 - \| \bA\x_{t+1} + \bB\z_{t+1} - \c \|_2^2  \big )~.
\end{align}
Plugging into \myref{eq:pos} and rearranging the terms complete the proof.
\qed

The above two lemmas together shows that
\begin{align}\label{eq:Rmono}
R(t+1,t+1)\leq R(t+1,t)\leq R(t,t) \leq R(t,t-1)~,
\end{align}
meaning $R(s,t)$ is monotonically non-increasing.
The following lemma shows $R(t+1,t)$ is bounded by step differences of a telescoping series of $\y$ and $\z$.
\begin{lem}\label{lem:opbd}
Let the sequences $\{\x_t, \z_t,\y_t\}$ be generated by ADM and $\{ \x^*, \z^*,\y^*\}$ satisfy the KKT conditions~\myref{eq:kkdx}-\myref{eq:kkdy}, then
\begin{align}\label{eq:lem_bconst}
R(t+1,t) \leq \|\bB\z^* - \bB\z_t\|_2^2 -  \|\bB\z^* - \bB\z_{t+1}\|_2^2 + \frac{1}{\rho^2} \left( \| \y^* - \y_t\|_2^2 - \| \y^* - \y_{t+1}\|_2^2 \right) ~.
\end{align}
\end{lem}
\proof
Assume $\{\x^*,\y^*\}$ satisfies~\myref{eq:kkdx}. Since $f$ is convex, then
\begin{align}
f(\x^*) -  f(\x_{t+1}) & \leq - \langle \bA^T\y^*, \x^* - \x_{t+1} \rangle  = - \langle \y^*, \bA\x^* - \bA\x_{t+1} \rangle~.
\end{align}
Similarly, for convex function $g$ and $\{\z^*,\y^*\}$ satisfies~\myref{eq:kkdz}, we have
\begin{align}
g(\z^*) - g(\z_{t+1}) & \leq - \langle \bB^T\y^*, \z^* - \z_{t+1} \rangle = - \langle \y^*, \bB\z^* - \bB\z_{t+1} \rangle ~.
\end{align}
Adding them together and using the fact that $\bA\x^* + \bB\z^* = \c$, we have
\begin{align}\label{eq:prop_bd}
& f(\x^*)+g(\z^*) - (f(\x_{t+1})+g(\z_{t+1})) \leq  \langle \y^*, \bA\x_{t+1} + \bB\z_{t+1} - \c \rangle~.
\end{align}
Adding \myref{eq:lem_bd1} and \myref{eq:prop_bd} together yields
\begin{align} \label{eq:cst_bd1}
 0 &\leq \frac{\rho}{2}  ( \|\bB\z^* - \bB\z_t\|_2^2 -  \|\bB\z^* - \bB\z_{t+1}\|_2^2 - \|\bA\x_{t+1} + \bB\z_t - \c\|_2^2 + \| \bA\x_{t+1} + \bB\z_{t+1} - \c\|_2^2 ) \nonumber \\
& + \langle \y^* -  \y_{t+1}, \bA\x_{t+1} + \bB\z_{t+1} - \c \rangle~.
\end{align}
The last term can be rewritten as
\begin{align}
& \langle \y^* - \y_{t+1}, \bA\x_{t+1} + \bB\z_{t+1} - \c \rangle = \frac{1}{\rho}\langle \y^* - \y_{t+1}, \y_{t+1} - \y_t \rangle \nonumber \\
& = -\frac{1}{2\rho}\left( \| \y^* - \y_t\|_2^2 - \| \y^* - \y_{t+1}\|_2^2 - \| \y_{t+1} - \y_t\|_2^2 \right)~.
\end{align}
Substituting it into \myref{eq:cst_bd1} and rearranging the terms gives
\begin{align}
& \|\bB\z^* - \bB\z_t\|_2^2 -  \|\bB\z^* - \bB\z_{t+1}\|_2^2 + \frac{1}{\rho^2} \left( \| \y^* - \y_t\|_2^2 - \| \y^* - \y_{t+1}\|_2^2 \right) \nonumber \\
&\geq \|\bA\x_{t+1} + \bB\z_t - \c\|_2^2 + \| \bA\x_{t+1} + \bB\z_{t+1} - \c\|_2^2 - \frac{1}{\rho^2} \| \y_{t+1} - \y_t\|_2^2 \nonumber \\
& = \|\bA\x_{t+1} + \bB\z_t - \c\|_2^2~,
\end{align}
which completes the proof.
\qed

Now, we are ready to show that the optimality conditions have a $O(1/T)$ convergence rate.
\begin{thm}\label{thm:cst_cr}
Let the sequences $\{\x_t, \z_t,\y_t\}$ be generated by ADM. For any $\x^*,\z^*$ satisfying $\bA\x^* + \bB\z^* = \c$, for any $T$, we have
\begin{align}
R(T,T) \leq R(T,T-1) \leq \frac{\lambda_{\max}^{\bB}D_\z^2 + D_\y^2/\rho^2}{T}~, \label{eq:thm_bcst}
\end{align}
where $R(T,T) = \|\bA\x_T + \bB\z_T - \c \|_2^2 + \|\bB\z_T - \bB\z_{T-1}\|_2^2$.
\label{thm:bcst}
\end{thm}
\proof
Since $\|\bA\x_{t+1} + \bB\z_{t} - \c\|_2^2$ is monotonically non-increasing, we have
\begin{align}
& TR(T,T-1) = T\|\bA\x_{T} + \bB\z_{T-1} - \c\|_2^2 \leq \sum_{t=0}^{T-1} \|\bA\x_{t+1} + \bB\z_{t} - \c\|_2^2 \nonumber \\
& \leq \|\bB\z^* - \bB\z_0\|_2^2 -  \|\bB\z^* - \bB\z_{T}\|_2^2 + \frac{1}{\rho^2} (\| \y^* - \y_0\|_2^2 - \| \y^* - \y_{T}\|_2^2) \nonumber \\
& \leq \|\bB\z^* - \bB\z_0\|_2^2 + \frac{1}{\rho^2}\| \y^* - \y_0\|_2^2~.
\end{align}
Divide both sides by $T$. Letting Assumption~\ref{asm:adm} hold and using Lemma \ref{lem:pdr} yield \myref{eq:thm_bcst}.
\qed


Results similar to Lemma \ref{lem:pdr} and \ref{lem:opbd} have appeared in~\cite{boyd10}, but Lemma \ref{lem:flbd} is new. The monotonicity and $O(1/T)$ convergence rate for optimality conditions have also been shown in ~\cite{he12:cst}, but our proof is different and self-contained.

\subsection{Rate of Convergence of ADM based on Variational Inequality}
We now prove the $O(1/T)$ convergence rate for ADM using a variational inequality (VI) based on the Lagrangian given in \myref{eq:lag1}. In this section, we need the following assumption~\cite{Bertsekas99,Bert96:bklag}:
\begin{asm}\label{asm:admvi}
$\y$ is bounded in $\R^m$ and $\| \y \|_2 \leq D$, i.e., $\y \in \mathcal{Y} \subset \R^m$ and $\mathcal{Y}$ is a bounded set.
\end{asm}
Let $\Omega = \cX \times \cZ \times \mathcal{Y}$, where $\cX$ and $\cZ$ are defined in~\myref{eq:admm_pm}. Any $\w^*=(\x^*,\z^*,\y^*) \in \Omega$ solves the original problem in~\myref{eq:badm_pm} optimally if it satisfies the following variational inequality~\cite{fapa03,Nemi:VI,he12:vi}:
\beq
\forall \w \in \Omega~, ~~~h(\w) - h(\w^*)+ \langle \w - \w^*, F(\w^*) \rangle \geq 0~,
\eeq
where $h(\w) = f(\x) + g(\z)$ and
\begin{align*}
F(\w) = \left [ \begin{array}{c}
\bA^T\y \\
\bB^T\y \\
-(\bA \x + \bB \z - \c)
\end{array}
\right ] =
\left [ \begin{array}{ccc}
\mathbf{0} & \mathbf{0} & \bA^T \\
\mathbf{0} & \mathbf{0} & \bB^T \\
-\bA & -\bB & \mathbf{0}
\end{array}
\right ]\w +
\left [ \begin{array}{c}
\mathbf{0} \\
\mathbf{0} \\
\c
\end{array}
\right ] = \bM\w + \q~
\end{align*}
is the gradient of the last term of the Lagrangian. $\bM$ is an anti-symmetric matrix and $\w^T\bM\w = 0$. Then, $\tilde{\w}=(\tilde{\x},\tilde{\z},\tilde{\y})$ approximately solves the problem with accuracy $\epsilon$ if it satisfies 
\beq
\forall \w \in \Omega~, ~~~ h(\tilde{\w}) - h(\w) + \langle \tilde{\w} - \w, F(\tilde{\w}) \rangle \leq \epsilon~.
\eeq
We show that after $T$ iterations, the average $\bar{\w}_T = \frac{1}{T} \sum_{t=1}^T \w_{t}$, where $\w_t = (\x_t,\z_t,\y_t)$ are from \myref{eq:admm_x}-\myref{eq:admm_y}, satisfies the above inequality with $\epsilon = O(1/T)$.
\begin{thm}
Let $\bar{\w}_T = \frac{1}{T} \sum_{t=1}^T \w_{t}$, where $\w_{t} = (\x_{t},\z_{t},\y_{t})$ from \myref{eq:admm_x}-\myref{eq:admm_y}. Let Assumption \ref{asm:adm} and \ref{asm:admvi} hold, then
\begin{equation*}
\forall \w \in \Omega, ~~~ h(\bar{\w}_T) - h(\w) + \langle \bar{\w}_T - \w , F(\bar{\w}_T) \rangle \leq \frac{L}{T}~.
\end{equation*}
where $L = \frac{\rho}{2} \| \bA\x - \c \|_2^2  + \frac{1}{2\rho}\| \y \|^2$.
\end{thm}
\proof Considering $f(\x)$ is a convex function and its subgradient is given in \myref{eq:fg2}, $\forall \x \in \cX$,
\begin{equation*}
f(\x_{t+1}) - f(\x) \leq - \langle \bA^T( \y_{t+1} + \rho (\bB \z_t - \bB\z_{t+1})), \x_{t+1} -\x \rangle ~.
\end{equation*}
Rearranging the terms gives
\begin{align}
& f(\x_{t+1})  - f(\x) + \langle \x_{t+1} - \x, \bA^T \y_{t+1} \rangle \leq  \rho \langle \bA \x - \bA \x_{t+1}, \bB \z_t - \bB \z_{t+1} \rangle~.
\label{eq:xvii}
\end{align}
Using the subgradient of $g$ given in \myref{eq:gy}, we have $\forall \z \in \cZ$
\begin{align}
g(\z_{t+1}) - g(\z) + \langle \z_{t+1} - \z, \bB^T \y_{t+1} \rangle \leq 0~.
\label{eq:zvii}
\end{align}
Adding \myref{eq:xvii} and \myref{eq:zvii} and denoting $h(\w) = f(\x) + g(\z)$, we have $\forall \w \in \Omega$
\begin{align}\label{eq:1step}
h&(\w_{t+1}) - h(\w) + \langle \w_{t+1} - \w, F(\w_{t+1}) \rangle  \\
&\leq ~ \rho \langle \bA \x - \bA \x_{t+1}, \bB \z_t - \bB \z_{t+1} \rangle
+ \frac{1}{\rho}\langle \y_{t+1} - \y, -(\bA\x_{t+1} + \bB\z_{t+1} -\c) \rangle \nonumber \\
&= ~ \rho \langle \bA \x - \bA \x_{t+1}, \bB \z_t - \bB \z_{t+1} \rangle
+ \frac{1}{\rho}\langle \y - \y_{t+1}, \y_{t+1} - \y_{t} \rangle~.\nonumber
\end{align}
The first term can be rewritten as
\begin{align}
2\langle & \bA \x - \bA \x_{t+1}, \bB \z_t - \bB \z_{t+1} \rangle \label{eq:xfst} \\
& = 2\langle \bA \x - \c - (\bA \x_{t+1} - \c), \bB \z_t - \bB \z_{t+1} \rangle \nonumber \\
& = \| \bA \x + \bB \z_{t} - \c\|^2 - \| \bA \x + \bB \z_{t+1} -\c \|^2 + \| \bA \x_{t+1} + \bB \z_{t+1} - \c \|^2 - \| \bA \x_{t+1} + \bB \z_{t} - \c \|^2~, \nonumber
\end{align}
where the last equality uses~\myref{eq:four}.
The second term in \myref{eq:1step} is equivalent to
\begin{align}\label{eq:ysec}
&2\langle \y - \y_{t+1}, \y_{t+1} - \y_t\rangle  =  \| \y - \y_{t} \|^2 - \| \y - \y_{t+1} \|^2 - \| \y_t - \y_{t+1} \|^2 ~,
\end{align}
which uses~\myref{eq:three}.
Substituting \myref{eq:xfst} and \myref{eq:ysec} into \myref{eq:1step} and using \myref{eq:admm_y}, we have
\begin{align}
h&(\w_{t+1}) - h(\w) + \langle \w_{t+1} - \w, F(\w_{t+1}) \rangle  \nonumber \\
& \leq \frac{\rho}{2}(\| \bA \x + \bB \z_{t} - \c\|^2 - \| \bA \x + \bB \z_{t+1} -\c \|^2) + \frac{1}{2\rho}(\| \y - \y_{t} \|^2 - \| \y - \y_{t+1} \|^2)~.
\end{align}
Summing over $t$ from $0$ to $T-1$, we have the following telescoping sum
\begin{align}\label{eq:sum1}
\sum_{t=1}^{T} \left [ h(\w_{t}) - h(\w) + \langle \w_{t} - \w,F(\w_{t})  \rangle\right ]  \leq  L~,
\end{align}
where the constant
$L = \frac{\rho}{2} \| \bA\x - \c \|_2^2  + \frac{1}{2\rho}\| \y \|^2$.
Recall that $h(\tilde{w})$ is a convex function of $\tilde{w}$. Further, from the definition of $F(\tilde{\w})$, we have
\begin{align}
\langle \tilde{\w} - \w, F(\tilde{\w}) \rangle = \langle \tilde{\w} - \w, \bM\tilde{\w} + \q \rangle =  - \langle \w, \bM\tilde{\w} \rangle + \langle \tilde{\w} - \w, \q \rangle~,
\end{align}
which is a linear function of $\tilde{\w}$. Dividing both sides of \myref{eq:sum1} by $T$, recalling that $\bar{\w}_T = \frac{1}{T} \sum_{t=1}^T \w_{t}$, and using Jensen's inequality, we have
\begin{align*}
\begin{split}
&h(\bar{\w}_T) - h(\w) + \langle \bar{\w}_T - \w, F(\bar{\w}_T) \rangle   \\
&\leq \frac{1}{T} \sum_{t=1}^{T} h(\w_{t}) - h(\w) + \frac{1}{T} \sum_{t=1}^T \langle \w_{t} - \w, F(\w_{t}) \rangle  \\
&\leq \frac{L}{T} = O\left(\frac{1}{T}\right)~,
\end{split}
\end{align*}
which establishes convergence rate for ADM.
\qed

The bound requires $\x$ and $\y$ to be bounded. In general,  $L$ is larger compard to the results in Theorem~\ref{thm:bobj} and~\ref{thm:cst_cr}. According to~\myref{eq:admm_x}, 
\begin{align}\label{eq:yT}
\rho\sum_{t=0}^{T-1} (\bA\x_{t+1} + \bB\z_{t+1} - \c) = \sum_{t=0}^{T-1}(\y_{t+1} - \y_t) = \y_T - \y_0 = \y_T~,
\end{align} 
meaning $\y_T$ is the sum of all past residuls of constraint violation and thus $\| \y \|_2$ is large.  \cite{he12:vi} also shows a similar result based on an auxiliary sequence $\{\x_{t+1},\z_{t+1},\tilde{\y}_{t+1} = \y_t + \rho(\bA\x_{t+1} + \bA\z_t - \c)\}$ instead of the sequence $\{\x_{t+1},\z_{t+1},\y_{t+1}\}$ generated by ADM. Compared to their proof, our proof is arguably simple and easier to understand. In fact, their proof is based on weak VI~\cite{Nemi:VI,David00:VI,fapa03}, while our proof is based on strong VI~\cite{Nemi:VI,David00:VI,fapa03}. According to Minty's lemma~\cite{David00:VI,fapa03}, they are equivalent if the solution set $\Omega$ is closed bounded and VI operator $F$ is continuous and monotone. 

\section{Online Alternating Direction Method}\label{sec:OADM}
In this section, we extend ADM to the online learning setting. Specifically, we focus on using online ADM (OADM) to solve the problem \myref{eq:admm_pm}. For our analysis, $\bA$ and $\bB$ are assumed to be fixed. At round $t$, we consider solving the following regularized optimization problem:
\begin{align}\label{eq:admm1}
\x_{t+1} = \underset{\bA\x + \bB \z = \c}{\argmin}  f_t(\x) + g(\z) +  \eta B_{\phi}(\x,\x_t)~,
\end{align}
where $\eta \geq 0$ is a learning rate and $B_{\phi}(\x,\x_t)$ is a Bregman divergence~\cite{bane05:bregman,ceze98}.

Let $\phi:\Omega\rightarrow\R$ be a continuously differentiable and strictly convex function. Denote $\nabla\phi(\y)$ as the gradient of $\phi$ at $\y$. The Bregman divergence $B_{\phi}: \Omega\times \ri(\Omega) \rightarrow \R_+$ is defined as
\begin{align*}
B_{\phi}(\x,\y) = \phi(\x) - \phi(\y) - \langle \nabla\phi(\y), \x - \y \rangle~.
\end{align*}

Two widely used examples are squared Euclidian distance $B_{\phi}(\x,\y) = \frac{1}{2}\|\x - \y\|_2^2$ and KL divergence $B_{\phi}(\x,\y) = \sum_{i=1}^n x_i \log \frac{x_i}{y_i}$.


If the problem~\myref{eq:admm1} is solved exactly in every step, standard analysis techniques~\cite{haak07} can be suitably adopted to obtain sublinear regret bounds. While \myref{eq:admm1} can be solved by batch ADM, we essentially obtain a double loop algorithm where the function $f_t$ changes in the outer loop and the inner loop runs ADM iteratively till convergence so that the constraint are satisfied. Note that existing online methods, such as projected gradient descent and variants~\cite{haak07,Duchi10_comid} do assume a black-box approach for projecting onto the feasible set, which for linear constraint may require
iterative cyclic projections~\cite{ceze98}.


 For our analysis, instead of requiring the equality constraint to be satisfied at each time $t$, we only require the equality constraint to be satisfied in the long run, with a notion of regret associated with constraint. In particular, we consider the following constrained cumulative regret for the online learning problem:
\begin{align}\label{eq:cstregret}
& \sum_{t=1}^T f_t(\x_t) + g(\z_t) - \min_{\bA\x + \bB\z = \c} \sum_{t=1}^T f_t(\x) + g(\z) \nonumber \\
& \text{s.t.} \quad \sum_{t=1}^T \| \bA\x_t + \bB\z_t - \c \|_2^2 = o(T)~,
\end{align}
where the cumulative constraint violation is sublinear in $T$. The goal is to design a single-loop algorithm for~\myref{eq:cstregret}, which has sublinear regret in both the objective and the constraint violation.

The augmented Lagrangian of \myref{eq:admm1} at time $t$ is
\begin{align}\label{eq:lag}
L_{\rho}^{t}(\x,\y,\z) = & f_t(\x) + g(\z) + \langle \y, \bA\x + \bB\z - \c \rangle  + \eta B_{\phi}(\x,\x_t) + \frac{\rho}{2} \| \bA\x + \bB\z - \c \|^2~.
\end{align}
At time $t$, OADM (\textbf{Algorithm \ref{alg:oadm}}) consists of just one pass through the following three update steps:
\begin{align}
\x_{t+1} &= \underset{\x}{\argmin}~ \{f_t(\x) + \langle \y_t, \bA\x + \bB\z_t - \c \rangle + \frac{\rho}{2} \| \bA\x + \bB\z_t -\c \|^2 + \eta B_{\phi}(\x,\x_t)\} ~, \label{eq:oadmm1_x}\\
\z_{t+1} & = \underset{\z}{\argmin}~ \{g(\z) + \langle \y_t, \bA\x_{t+1} + \bB\z - \c \rangle + \frac{\rho}{2} \| \bA\x_{t+1} + \bB\z - \c \|^2\}  ~, \label{eq:oadmm1_z} \\
\y_{t+1} & = \y_t + \rho(\bA\x_{t+1} + \bB\z_{t+1} - \c) ~.\label{eq:oadmm1_y}
\end{align}




Operationally, in round $t$, the algorithm presents a solution $\{\x_{t},\z_{t}\}$ as well as $\y_{t}$. Then, nature reveals  function $f_t$ and we encounter two types of losses. The first type is the traditional loss measured by $f_t(\x_t) + g(\z_t)$, with corresponding cumulative regret
\begin{equation}\label{eq:rgt1}
R_1(T) = \sum_{t=1}^T f_t(\x_t) + g(\z_t) - \min_{\bA\x + \bB\z = \c} \sum_{t=1}^T f_t(\x) + g(\z)~.
\end{equation}
The second type is the residual of constraint violation, i.e., $\| \bA\x_t + \bB \z_t - \c\|^2$. As the updates include the primal and dual variables, in line with batch ADM, we use the following cumulative regret for constraint violation:
\begin{align}\label{eq:Rc}
R^c(T) = \sum_{t=1}^T \| \bA\x_{t+1} + \bB\z_{t+1} - \c \|_2^2 + \| \bB\z_{t+1} - \bB\z_t \|_2^2~.
\end{align}
The goal is to establish sublinear regret bounds for both the objective and constraint violation.

\begin{algorithm*}[tb]
\caption{Online Alternating Direction Method (OADM)}
\label{alg:oadm}
\begin{algorithmic}[1]
\STATE {\bfseries Input:} $f_{t}(\x) + g(\z), \bA,\bB,\c, \rho, \eta, \phi(\x)$
\STATE {\bfseries Initialization:} $\x_1,\z_1,\u_1 = \mathbf{0}$
\FOR{$t=1 \text{ to } T$}
\STATE $\x_{t+1} = \argmin_{\x}~ f_t(\x) + \langle \y_t, \bA\x + \bB\z_t - \c \rangle + \frac{\rho}{2} \| \bA\x + \bB\z_t -\c \|^2 + \eta B_{\phi}(\x,\x_t)$ ~,\\
\STATE $\z_{t+1} = \argmin_{\z}~ g(\z) + \langle \y_t, \bA\x_{t+1} + \bB\z - \c \rangle + \frac{\rho}{2} \| \bA\x_{t+1} + \bB\z - \c \|^2$  ~,  \\
\STATE $\y_{t+1} = \y_t + \rho(\bA\x_{t+1} + \bB\z_{t+1} - \c)$ ~.
\STATE Receive a cost function $f_{t+1}$ and incur loss $f_{t+1}(\x_{t+1}) + g(\z_{t+1})$ and constraint violation $\| \bA\x_{t+1} + \bB\z_{t+1} - \c\|_2^2$;
\ENDFOR
\end{algorithmic}
\end{algorithm*}

The OADM updates~\myref{eq:oadmm1_x}-\myref{eq:oadmm1_z} are similar as ADM updates~\myref{eq:admm_x}-\myref{eq:admm_z} except the $\x$ update in OADM uses a time varying function $f_t$ and an additional Bregman divergence, which is the first scenario where the regret bounds of $R_1$~\myref{eq:rgt1} and $R^c$~\myref{eq:Rc} will be presented in Section 4. 
We also consider another scenario, where $\eta = 0$ in \myref{eq:oadmm1_x} and thus the Bregman divergence is eliminated and only the quadratic penalty term is involved in the $\x$-update. $\x_{t+1}$ is kept close to $\x_t$ indirectly through the quadratic penalty term at $\z_t$.
Instead of using $\{\x_t,\z_t\}$ as the solution at round $t$, we use a solution $\{\hat{\x}_t,\z_t \}$ based on $\z_t$ such that $\bA\hat{\x}_t + \bB \z_t = \c$. While $\{\hat{\x}_t,\z_t\}$ satisfies the constraint by design, the goal is to establish sublinear regret of the objective $f_t(\hat{\x}_t) + g(\z_t)$, i.e.,
\begin{equation}\label{eq:rgt2}
R_2(T) = \sum_{t=1}^T f_t(\hat{\x}_t) + g(\z_t) - \min_{\bA\x + \bB\z = \c} \sum_{t=1}^T f_t(\x) + g(\z)~.
\end{equation}
The sublinear regret of constraint violation for the true $\{\x_t,\z_t\}$ defined in~\myref{eq:Rc} should still be achieved.
The regret bounds for OADM in the two scenarios are summarized in Table \ref{tbl:rgt_oadm}.

\begin{table}[tb]
\centering
\begin{tabular}{|c|c|c|c|c|}
\hline
\multirow{2}{*}{Regret bounds} &\multicolumn{2}{|c|}{ $\eta > 0$}&\multicolumn{2}{|c|}{$\eta = 0$} \\
\cline{2-5}
& $R_1$ & $R^c$ & $R_2$ & $R^c$ \\
\hline
general convex & $O(\sqrt{T})$ & $O(\sqrt{T})$ & $O(\sqrt{T})$ & $O(\sqrt{T})$ \\
\hline
strongly convex & $O(\log{T})$ & $O(\log{T})$ & $O(\log{T})$ & $O(\log{T})$ \\
\hline
\end{tabular}
\caption{Regret Bounds for Online Alternating Direction Method}
\label{tbl:rgt_oadm}
\end{table}

Before getting into the regret analysis, we discuss some example problems which can be solved using OADM. Like FOBOS and RDA, OADM can deal with machine learning problems where $f_t$ is a loss function and $g$ is a regularizer, e.g., generalized lasso and group lasso ~\cite{boyd10,Tibshirani96,xiao10} using $\ell_1$ or mixed norm, or an indicator function of a convex set. OADM can also be used to solve the batch optimization problems mentioned in Section 1, including linear programs, e.g.,
MAP LP relaxation~\cite{Meshi10} and LP decoding~\cite{Barman11}, and non-smooth optimization, e.g. robust PCA~\cite{cand09:rpca,Lin_Ma09}. Another promising scenario for OADM is consensus optimization~\cite{boyd10} where distributed local variables are updated separately and reach a global consensus in the long run. More examples can be found in~\cite{boyd10} and references therein.

In the sequel, we need the following assumptions:
\begin{asm}\label{asm:oadm}
\quad

(a) For a $p$-norm $\|\cdot\|_p$, the dual norm of subgradient of $f_t(\x)$ is bounded by $G_f$, i.e., $\| \nabla f'_t(\x)\|_q \leq G_f$, where $f'_t(\x) \in\partial f_t(\x), \forall \x\in\cX$ and $\frac{1}{p} + \frac{1}{q} = 1$.

(b) The Bregman divergence $B_{\phi}$ is defined on an $\alpha$-strongly convex function $\phi$ with respect to a $p$-norm $\|\cdot\|_p$, i.e., $B_{\phi}(\u,\v) \geq \frac{\alpha}{2} \|\x - \y \|_p^2$
where $\alpha > 0$. 


(c) $\x_1 = \mathbf{0}, \y_1 = \mathbf{0}, \z_1 = \mathbf{0}$. For any $\x^*, \z^*$ satisfying $\bA\x^*+\bB\z^* = \c$, $B_{\phi}(\x^*,\x_1) \leq D_{\x}^2, \| \z^* - \z_1 \|_2 \leq D_{\z}$.

(d) $g(\z_1) = 0$ and $g(\z) \geq 0$.

(e) For any $t$, $f_t(\x_{t+1}) + g(\z_{t+1}) - (f_t(\z^*) + g(\z^*)) \geq -F$, where $F$ is a positive constant.
\end{asm}

In Assumption~\ref{asm:oadm}, (a) and (b) are in general required in the online learning setting~\cite{Zinkevich03,duchi09,xiao10}. (c) and (d) are simply for the ease of exposition of regret bounds and is commonly assumed for composite objective~\cite{duchi09,xiao10}, e.g., $g$ is a regularizer in machine learning. We may assume the convex sets of $\x$ and $\z$ are bounded~\cite{Zinkevich03,haak07} in (c).  To obtain a sublinear regret bound for constraint violation, we need (e), which is true if functions are bounded from below or Lipschitz continuous in the convex set ~\cite{Majin11}.

%
%



\section{Regret Analysis for OADM}\label{sec:oadmm1}
We consider two types of regret in OADM. The first type is the regret of the objective based on splitting variables, i.e., $R_1$ defined in~\myref{eq:rgt1}.
Aside from using splitting variables, $R_1$ is the standard regret in the online learning setting.
The second is the regret of the constraint violation $R^c$ defined in \myref{eq:Rc}. We establish sublinear regret bounds for several cases whether $f_t$ and $g$ are strongly convex or not.



\subsection{General Convex Functions}

The following establishes the regret bounds for OADM for general convex functions.

\begin{thm}\label{thm:oadmm1_rgtbd}
Let the sequences $\{\x_t, \z_t,\y_t\}$ be generated by OADM~\myref{eq:oadmm1_x}-\myref{eq:oadmm1_y} and let Assumption~\ref{asm:oadm} hold. For any $\x^*,\z^*$ satisfying $\bA\x^* + \bB\z^* = \c$, setting $\eta = \frac{G_f\sqrt{T}}{D_\x \sqrt{2\alpha}}$ and $\rho = \sqrt{T}$,  we have
\begin{align}
&R_1(T)
\leq \frac{\lambda_{\max}^{\bB}D_{\z}^2\sqrt{T}}{2} + \frac{\sqrt{2}G_fD_\x\sqrt{T}}{\sqrt{\alpha}} \label{eq:oadmm1_bd} ~, \\
&R^c(T) \leq \lambda_{\max}^{\bB}D_\z^2 +\frac{2\sqrt{2}D_{\x}G_f}{\sqrt{\alpha}} + 2F\sqrt{T} \label{eq:oadmm1_cst2}~.
\end{align}
\end{thm}
\proof
Since $\x_{t+1}$ minimizes \myref{eq:oadmm1_x}, we have
\begin{align}
0 \in \partial f_t(\x_{t+1}) + \bA^T\y_t + \rho\bA^T(\bA\x_{t+1} - \bB\z_t - \c) + \eta(\nabla \phi(\x_{t+1}) - \nabla \phi(\x_t)) ~.
\end{align}
Rearranging the terms and using~\myref{eq:oadmm1_y} give the subgradient of $f_t(\x_{t+1})$, 
\begin{align}\label{eq:oadmm1_fg}
-\bA^T(\y_{t+1} + \rho(\bB\z_t - \bB\z_{t+1})) - \eta(\nabla \phi(\x_{t+1}) - \nabla \phi(\x_t)) \in \partial f_t(\x_{t+1})
\end{align}
Compared to~\myref{eq:fg2} in Lemma~\ref{lem:sg}, the additional terms introduced by Bregman divergence are included in the subgradient.  Therefore, replacing $f$ by $f_t$ in Lemma \ref{lem:bd1} and adding the terms $- \eta(\nabla \phi(\x_{t+1}) - \nabla \phi(\x_t))$, we have
\begin{align}\label{eq:oadmm1_bd1}
&f_t(\x_{t+1}) + g(\z_{t+1}) - (f_t(\x^*) + g(\z^*)) \nonumber \\
& \leq \frac{1}{2\rho}(\| \y_{t} \|_2^2 - \| \y_{t+1} \|_2^2)
- \frac{\rho}{2} \|\bA\x_{t+1} + \bB\z_t - \c\|_2^2 + \frac{\rho}{2}  ( \|\bB\z^* - \bB\z_t\|_2^2 -  \|\bB\z^* - \bB\z_{t+1}\|_2^2)  \nonumber \\
&   - \eta \langle \nabla \phi(\x_{t+1}) - \nabla \phi(\x_t), \x_{t+1} - \x^* \rangle~.
\end{align}
Using the three point property of Bregman divergence, the last term can be written as 
\begin{align}\label{eq:bgmthree}
 - \langle \nabla \phi(\x_{t+1}) - \nabla \phi(\x_t), \x_{t+1} - \x^* \rangle = B_{\phi}(\x^*,\x_t) - B_{\phi}(\x^*, \x_{t+1}) - B_{\phi}(\x_{t+1} , \x_t)~.
\end{align}
Let $f_t'(\x_{t}) \in \partial f_t(\x_{t})$. According to the Fenchel-Young's inequality~\cite{rock70}, i.e., $2 |\langle \x, \y \rangle| \leq \| \x\|_q^2 + \| \y \|_p^2$, we have
\begin{align}\label{eq:oadmm1_fb0}
f_t(\x_{t}) - f_t(\x_{t+1}) & \leq \langle f_t'(\x_{t}), \x_t - \x_{t+1} \rangle =  \langle \frac{1}{\sqrt{\alpha\eta}}f_t'(\x_{t}), \sqrt{\alpha\eta} (\x_t - \x_{t+1}) \rangle \nonumber \\
& \leq \frac{1}{2\alpha\eta} \| f_t'(\x_{t}) \|_q^2 + \frac{\alpha\eta}{2} \|\x_t - \x_{t+1}\|_p^2 ~.
\end{align}
Recalling $B_{\phi}(\x_{t+1} , \x_t) \geq \frac{\alpha}{2} \|\x_t - \x_{t+1}\|_p^2$ and combining~\myref{eq:oadmm1_bd1}-\myref{eq:oadmm1_fb0}, we have
\begin{align}\label{eq:oadm1_r0}
&f_t(\x_{t}) + g(\z_{t+1}) - (f_t(\x^*) + g(\z^*)) \nonumber \\
& \leq \frac{1}{2\rho}(\| \y_{t} \|_2^2 - \| \y_{t+1} \|_2^2)
- \frac{\rho}{2} \|\bA\x_{t+1} + \bB\z_t - \c\|_2^2
+ \frac{\rho}{2}  ( \|\bB\z^* - \bB\z_t\|_2^2 -  \|\bB\z^* - \bB\z_{t+1}\|_2^2) \nonumber \\
& \quad + \frac{1}{2\alpha\eta} \| f_t'(\x_{t}) \|_q^2  + \eta (B_{\phi}(\x^*,\x_t) - B_{\phi}(\x^*, \x_{t+1})) ~.
\end{align}
From Assumption~\ref{asm:oadm},  $g(\z) \geq 0$ and $g(\z_1) = 0$ for $\z_1 = \mathbf{0}$, $R_1(T)$ is bounded as follows :
\begin{align}\label{eq:oadm1_r01}
R_1(T)& =  \sum_{t=1}^T f_t(\x_{t}) + g(\z_{t+1}) - (f_t(\x^*) + g(\z^*)) + g(\z_1) - g(\z_{T+1}) \nonumber \\
& \leq  \frac{1}{2\rho}(\| \y_{1} \|_2^2 - \| \y_{T+1} \|_2^2)
+ \frac{\rho}{2}  ( \|\bB\z^* - \bB\z_1\|_2^2 -  \|\bB\z^* - \bB\z_{T+1}\|_2^2) \nonumber \\
&
\quad + \eta (B_{\phi}(\x^*,\x_1) - B_{\phi}(\x^*, \x_{T+1}) ) +\frac{1}{2\alpha\eta} \sum_{t=1}^T \| f_t'(\x_{t}) \|_q^2 \nonumber \\
& \leq \frac{\lambda_{\max}^{\bB}D_{\z}^2\rho}{2} + \eta D_{\x}^2 + \frac{G_f^2T}{2\alpha \eta} ~.
\end{align}
Setting $\eta = \frac{G_f\sqrt{T}}{D_\x \sqrt{2\alpha}}$ and $\rho = \sqrt{T}$ yields \myref{eq:oadmm1_bd}.

Now we prove \myref{eq:oadmm1_cst2}. Rearranging the terms in \myref{eq:oadmm1_bd1}, we have
\begin{align}\label{eq:oadm1_cstone}
\|\bA\x_{t+1} + \bB\z_t - \c\|_2^2 & \leq \frac{2F}{\rho} + \frac{1}{\rho^2}(\| \y_{t} \|_2^2 - \| \y_{t+1} \|_2^2) +  ( \|\bB\z^* - \bB\z_t\|_2^2 -  \|\bB\z^* - \bB\z_{t+1}\|_2^2)  \nonumber \\
&\quad + \frac{2\eta}{\rho} (B_{\phi}(\x^*,\x_t) - B_{\phi}(\x^*, \x_{t+1}) - B_{\phi}(\x_{t+1} , \x_t)) ~.
\end{align}
Letting Assumption~\ref{asm:oadm} hold and summing over $t$ from 1 to $T$, we have
\begin{align}\label{eq:oadm1_cstsum}
& \sum_{t=1}^T \|\bA\x_{t+1} + \bB\z_t - \c\|_2^2 \nonumber \\
& \leq \frac{2FT}{\rho} + \frac{1}{\rho^2}(\| \y_{1} \|_2^2 - \| \y_{T+1} \|_2^2) +  ( \|\bB\z^* - \bB\z_1\|_2^2 -  \|\bB\z^* - \bB\z_{T+1}\|_2^2)  \nonumber \\
& \quad + \frac{2\eta}{\rho} (B_{\phi}(\x^*,\x_1) - B_{\phi}(\x^*, \x_{T+1})) \nonumber \\
& \leq \frac{2FT}{\rho} + \lambda_{\max}^{\bB}D_\z^2 + \frac{2\eta}{\rho}D_{\x}^2~.
\end{align}
Setting $\eta = \frac{G_f\sqrt{T}}{D_\x \sqrt{2\alpha}}$ and $\rho = \sqrt{T}$, we have \myref{eq:oadmm1_cst2} by using Lemma \ref{lem:pdr}.
\qed

Note the bounds are achieved without any explicit assumptions on $\bA,\bB,\c$.\footnote{We do assume that $\bA\x + \bB\z = \c$ is feasible.}  The subgradient of $f_t$ is required to be bounded, but the subgradient of $g$ is not necessarily bounded. Thus, the bounds hold for the case where $g$ is an indicator function of a convex set. Compared to regret bound for COMID which is $\frac{G_fD_\x\sqrt{T}}{\sqrt{\alpha}}$~\cite{Duchi10_comid}, the regret bound for the objective of ADMM has an additional term $\frac{\lambda_{\max}^{\bB}D_{\z}^2\sqrt{T}}{2}$ which is for the splitting variable $\z$. In addition to the $O(\sqrt{T})$ regret bound, OADM achieves the $O(\sqrt{T})$ bound for the constraint violation, which is not considered in the start-of-the-art online learning algorithms ~\cite{Duchi10_comid,duchi09,xiao10}, since they do not explicitly handle linear constraint of the form $\bA\x+\bB\z =\c$. In fact, the bound for constraint violation could be reduced to a constant if $\y_t$ is assumed to be bounded (see Assumption~\ref{asm:admvi}), which is shown in the following theorem.

\begin{thm} \label{thm:cstbd}
Let the sequences $\{\x_t, \z_t,\y_t\}$ be generated by OADM. Assume that $\|\y_t\|_2 \leq D$. Setting $\rho = \sqrt{T}$, then
\begin{align}
&\sum_{t=1}^T \| \bA\x_{t+1} + \bB\z_{t+1} - \c \|_2^2 \leq 4D^2~. \label{eq:oadmm1_cst1}
\end{align}
\end{thm}
\proof
According to \myref{eq:oadmm1_y}, we have
\begin{align}\label{eq:cstbd01}
&\| \bA\x_{t+1} + \bB\z_{t+1} - \c \|_2^2  = \| \frac{1}{\rho}(\y_{t+1} - \y_t) \|_2^2 \leq \frac{2}{\rho^2} (\| \y_{t+1} \|_2^2 + \| \y_t \|_2^2) \leq \frac{4D^2}{\rho^2}~.
\end{align}
Summing over $t$ from 1 to $T$ and setting $\rho = \sqrt{T}$ yield \myref{eq:oadmm1_cst1}.
\qed

\subsection{Strongly Convex Functions}
We assume both $f_t(\x)$ and $g$ are strongly convex.
Specifically, we assume $f_t(\x)$ is $\beta_1$-strongly convex with respect to a differentiable convex function $\phi$, i.e.,
\begin{align}
& f_t(\x^*) \geq  f_t(\x) + \langle f_t'(\x), \x^* - \x \rangle + \beta_1 B_{\phi} (\x^*,\x)~,\label{eq:fsc}
\end{align}
where $f_t'(\x)$ denotes the subgradient of $f_t$ at $\x$ and $\beta_1 > 0$. Assume $g$ is a $\beta_2$-strongly convex function, i.e.,
\begin{align}
& g(\z^*) \geq g(\z) + \langle g'(\z), \z^* - \z \rangle +  \frac{\beta_2}{2} \| \z^* -  \z \|_2^2~, \label{eq:gsc}
\end{align}
where $g'(\z)$ denotes the subgradient of $g$ at $\z$ and $\beta_2>0$.

Instead of using fixed $\rho$ and $\eta$, we allow them to change over time, i.e., $\rho_t$ and $\eta_t$, which is fairly standard in the proof of logarithmic regret bounds~\cite{haak07,duchi09,xiao10} where the curvature of a sequence of strongly convex functions $f_t$ is considered. The following theorem establishes logarithmic regret bounds for $R_1$ as well as $R^c$.


\begin{thm} \label{thm:oadm_logrgtbd1}
Let Assumption~\ref{asm:oadm} hold. Assume $f_t(\x)$ and $g$ are strongly convex given in \myref{eq:fsc} and \myref{eq:gsc}. Setting $\eta_t = \beta_1 t, \rho_t = \beta_2 t/\lambda_{\max}^{\bB}$, we have
\begin{align}
& R_1(T) \leq \frac{G_f^2}{2\alpha\beta_1} \log{(T+1)} + \frac{\beta_2 D_{\z}^2}{2} + \beta_1 D_{\x}^2~, \label{eq:logrgt11} \\
&R^c(T) \leq \frac{2F\lambda_{\max}^{\bB}}{\beta_2}\log(T+1) + \lambda_{\max}^{\bB}D_\z^2 + \frac{2\beta_1\lambda_{\max}^{\bB}D_{\x}^2}{\beta_2} \label{eq:logrgt11_cst}~.
\end{align}
\end{thm}
\proof
Assume $f_t(\x)$ and $g$ are strongly convex~\myref{eq:fsc}-\myref{eq:gsc}. Let $\x$ be $\x_{t+1}$ and $\z$ be $\z_{t+1}$ in~\myref{eq:fsc}-\myref{eq:gsc} respectively. Adding them together and rearranging the terms give
\begin{align}\label{eq:fgstrong0}
&f_t(\x_{t+1}) + g(\z_{t+1}) - (f_t(\x^*)+ g(\z^*)) \nonumber \\
&\leq \langle f_t'(\x_{t+1}), \x_{t+1} - \x^* \rangle - \beta_1 B_{\phi} (\x^*,\x_{t+1}) + \langle g'(\z_{t+1}), \z_{t+1} - \z^* \rangle -  \frac{\beta_2}{2} \| \z^* -  \z_{t+1} \|_2^2~.
\end{align}
Compared to the general convex case in Theorem~\ref{thm:oadmm1_rgtbd}, the right hand side has two additional strongly convex terms.  \myref{eq:fgstrong0} can be obtained by 
letting $\rho, \eta$ be $\rho_{t+1}, \eta_{t+1}$ respectively in~\myref{eq:oadmm1_bd1} and adding the two strongly convex term as follows:
\begin{align}\label{eq:logbdfg}
&f_t(\x_{t+1}) + g(\z_{t+1}) - (f_t(\x^*)+ g(\z^*)) \nonumber \\
& \leq \frac{1}{2\rho_{t+1}}(\| \y_{t} \|_2^2 - \| \y_{t+1} \|_2^2) - \frac{\rho_{t+1}}{2} \| \bA\x_{t+1} + \bB\z_{t} - \c\|_2^2 + \frac{\rho_{t+1}}{2} ( \|\bB\z^* - \bB\z_t\|_2^2 -  \|\bB\z^* - \bB\z_{t+1}\|_2^2) \nonumber \\
& +\eta_{t+1} (B_{\phi}(\x^*,\x_t) - B_{\phi}(\x^*, \x_{t+1}) - B_{\phi}(\x_{t+1},\x_t)) - \beta_1 B_{\phi}(\x^*, \x_{t+1}) - \frac{\beta_2}{2}\|\z^* - \z_{t+1}\|_2^2~.
\end{align}
Let $\eta$ be $\eta_{t+1}$ in~\myref{eq:oadmm1_fb0}. Adding to~\myref{eq:logbdfg} and ignoring the negative term $- \frac{\rho_{t+1}}{2} \| \bA\x_{t+1} + \bB\z_{t} - \c\|_2^2$, we have 
\begin{align}
&f_t(\x_{t}) + g(\z_{t+1}) - (f_t(\x^*)+ g(\z^*)) \nonumber \\
& \leq  \frac{1}{\eta_{t+1}}\| f_t'(\x_{t}) \|_*^2 + \frac{1}{2\rho_{t+1}}(\| \y_{t} \|_2^2 - \| \y_{t+1} \|_2^2) + \frac{\rho_{t+1}}{2} ( \|\bB\z^* - \bB\z_t\|_2^2 -  \|\bB\z^* - \bB\z_{t+1}\|_2^2) \nonumber \\
& - \frac{\beta_2}{2}\|\z^* - \z_{t+1}\|_2^2 +(\eta_{t+1} (B_{\phi}(\x^*,\x_t) - B_{\phi}(\x^*, \x_{t+1})) - \beta_1 B_{\phi}(\x^*, \x_{t+1}) ) - \eta_{t+1} B_{\phi}(\x_{t+1},\x_t)~.
\end{align}
Summing over $t$ from 1 to $T$, we have
\begin{align}\label{eq:R1_logrgt}
R_1(T) & \leq  \frac{1}{2\alpha} \sum_{t=1}^T  \frac{1}{\eta_{t+1}}\| f_t'(\x_{t}) \|_*^2 + \sum_{t=1}^T  \frac{1}{2\rho_{t+1}}(\| \y_{t} \|_2^2 - \| \y_{t+1} \|_2^2) \nonumber \\
& + \sum_{t=1}^T (\frac{\rho_{t+1}}{2}  ( \|\bB\z^* - \bB\z_t\|_2^2 -  \|\bB\z^* - \bB\z_{t+1}\|_2^2) - \frac{\beta_2}{2}\|\z^* - \z_{t+1}\|_2^2) \nonumber \\
& + \sum_{t=1}^T (\eta_{t+1} (B_{\phi}(\x^*,\x_t) - B_{\phi}(\x^*, \x_{t+1})) - \beta_1 B_{\phi}(\x^*, \x_{t+1}) )~.
\end{align}
Assuming $\rho_t$ is non-decreasing, we have
\begin{align}\label{eq:yrho}
\sum_{t=1}^T\frac{1}{2\rho_{t+1}}(\| \y_{t} \|_2^2 - \| \y_{t+1} \|_2^2) \leq \frac{1}{2\rho_{2}}\| \y_{1} \|_2^2 = 0~.
\end{align}
Using $\|\bB\z^* - \bB\z_{t+1}\|_2^2 \leq \lambda_{\max}^{\bB}\|\z^* - \z_{t+1}\|_2^2$ and setting $\rho_t = \beta_2 t/\lambda_{\max}^{\bB}$, we have
\begin{align}\label{eq:logrgt_z}
& \sum_{t=1}^T \left [  \rho_{t+1}  ( \|\bB\z^* - \bB\z_t\|_2^2 -  \|\bB\z^* - \bB\z_{t+1}\|_2^2 ) - \beta_2\|\z^* - \z_{t+1}\|_2^2  \right ]   \nonumber \\
& \leq \sum_{t=1}^T  \left [ \rho_{t+1}  ( \|\bB\z^* - \bB\z_t\|_2^2 -  \|\bB\z^* - \bB\z_{t+1}\|_2^2 ) - \frac{\beta_2}{\lambda_{\max}^{\bB}}\|\bB\z^* - \bB\z_{t+1}\|_2^2 \right ] \nonumber \\
& \leq \rho_2\|\bB\z^* - \bB\z_1\|_2^2 + \sum_{t=2}^T \|\bB\z^* - \bB\z_t\|_2^2 (\rho_{t+1} - \rho_{t} - \frac{\beta_2}{\lambda_{\max}^{\bB}}) \nonumber \\
& = 2\beta_2 D_{\z}^2~,
\end{align}
where the last equality uses the Assumption~\ref{asm:oadm}.  Similarly, 
setting $\eta_t = \beta_1 t$, the last term in \myref{eq:R1_logrgt} can be rewritten as
\begin{align}\label{eq:logrgt_phi}
&\sum_{t=1}^T \left[ \eta_{t+1} (B_{\phi}(\x^*,\x_t) - B_{\phi}(\x^*, \x_{t+1})) - \beta_1 B_{\phi}(\x^*, \x_{t+1}) \right] \nonumber \\
& = \eta_2 B_{\phi}(\x^*,\x_1) + \sum_{t=2}^T B_{\phi}(\x^*,\x_t)(\eta_{t+1} - \eta_{t} -\beta_1 ) -\eta_{T+1}B_{\phi}(\x^*,\x_{T+1}) - \beta_1 B_{\phi}(\x^*, \x_{T+1})\nonumber \\
& \leq \eta_2 B_{\phi}(\x^*,\x_1) + \sum_{t=2}^T B_{\phi}(\x^*,\x_t)(\eta_{t+1} - \eta_{t} -\beta_1 ) \nonumber \\
& = 2\beta_1D^2_\x~.
\end{align}
Setting $\rho_t = \beta_2 t/\lambda_{\max}^{\bB}, \eta_t = \beta_1 t$ and combining \myref{eq:R1_logrgt}, \myref{eq:yrho}, \myref{eq:logrgt_z} and \myref{eq:logrgt_phi}, we have
\begin{align}
& R_1(T) \leq \frac{G_f^2}{2\alpha} \sum_{t=1}^T \frac{1}{\beta_1 (t+1)} + \beta_2 D_{\z}^2 + 2\beta_1 D_{\x}^2~.
\end{align}
Applying $\sum_{t=1}^T \frac{1}{t+1} \leq \int_{t=0}^T \frac{1}{t+1}dt = \log(T+1)$ gives \myref{eq:logrgt11}.

Now we prove \myref{eq:logrgt11_cst}. Rearranging terms in \myref{eq:logbdfg}, we have
\begin{align}\label{eq:log11_cstone}
\|\bA\x_{t+1} + \bB\z_t - \c\|_2^2 & \leq \frac{2F}{\rho_{t+1}} + \frac{1}{\rho_{t+1}^2}(\| \y_{t} \|_2^2 - \| \y_{t+1} \|_2^2) +  ( \|\bB\z^* - \bB\z_t\|_2^2 -  \|\bB\z^* - \bB\z_{t+1}\|_2^2)  \nonumber \\
&+ \frac{2\eta_{t+1}}{\rho_{t+1}} (B_{\phi}(\x^*,\x_t) - B_{\phi}(\x^*, \x_{t+1}) - B_{\phi}(\x_{t+1} , \x_t)) ~.
\end{align}
Letting $\rho_t = \beta_2 t/\lambda_{\max}^{\bB}$ and $\eta_t = \beta_1 t$ and summing over $t$ from 0 to $T$, we have

\begin{align}
& \sum_{t=1}^T \|\bA\x_{t+1} + \bB\z_t - \c\|_2^2 \nonumber \\
& \leq 2F\sum_{t=1}^T\frac{1}{\rho_{t+1}} + \sum_{t=1}^T\frac{1}{\rho_{t+1}^2}(\| \y_{t} \|_2^2 - \| \y_{t+1} \|_2^2) +  ( \|\bB\z^* - \bB\z_0\|_2^2 -  \|\bB\z^* - \bB\z_{T+1}\|_2^2)  \nonumber \\
&+ \sum_{t=1}^T\frac{2\eta_{t+1}}{\rho_{t+1}} (B_{\phi}(\x^*,\x_t) - B_{\phi}(\x^*, \x_{t+1})) \nonumber \\
& \leq \frac{2F\lambda_{\max}^{\bB}\log(T+1)}{\beta_2} + \lambda_{\max}^{\bB}D_\z^2 + \frac{2\beta_1\lambda_{\max}^{\bB}D_{\x}^2}{\beta_2}~.
\end{align}
We use \myref{eq:yrho} in the last inequality. According to Lemma \ref{lem:pdr}, we have \myref{eq:logrgt11_cst}.
\qed


To guarantee logarithmic regret bounds for both objective and constraints violation, OADM requires both $f_t$ and $g$ to be strongly convex. FOBOS, COMID, and RDA only require $g$ to be strongly convex although they do not consider linear constraints explicitly. Further, the logarithmic regret bounds for the constraints violation could reduce to constant bound if assuming $\y_t$ is bounded.

\begin{thm}
Let the sequences $\{\x_t, \z_t,\y_t\}$ be generated by OADM and $\| \y_t \|_2 \leq D$. Setting $\rho_t = \beta_2 t/\lambda_{\max}^{\bB}$, then
\begin{align}
&\sum_{t=1}^T \| \bA\x_{t+1} + \bB\z_{t+1} - \c \|_2^2 \leq \frac{2\pi D^2{\lambda_{\max}^{\bB}}^2}{3\beta_2^2}~.
\end{align}
\end{thm}
\proof Replacing $\rho$ by $\rho_{t+1}$ in~\myref{eq:cstbd01} and summing over $t$ from 1 to $T$, we have
\begin{align}
\sum_{t=1}^T\| \bA\x_{t+1} + \bB\z_{t+1} - \c \|_2^2 \leq \sum_{t=1}^T\frac{4D^2}{\rho_{t+1}^2}~.
\end{align}
Setting $\rho_t = \beta_2 t/\lambda_{\max}^{\bB}$ and using $\sum_{t=1}^T \frac{1}{t^2} \leq \frac{\pi}{6}$ complete the proof.
\qed

\section{Regret Analysis for OADM with $\eta = 0$}\label{sec:oadmm2}
We analyze the regret bound when $\eta = 0$. In this case, OADM has the same updates as ADM except $f_t$ is changing over time. The $\x$-update only including the quadratic penalty term is easier to solve than the one with an additional Bregman divergence, particularly when the Bregman divergence is not a quadratic function.  Without a Bregman divergence to keep two consecutive iterates of $\x$ close, the quadratic penalty term is qualified for this task through variable $\z$. We consider $\z_t$ to be the key primal variable, and compute $\hat{\x}_t$ using $\z_t$ so that $\bA\hat{\x}_t + \bB\z_t = \c$. Therefore, we use the regret bound $R_2$ defined in~\myref{eq:rgt2}.
While $\{\hat{\x}_t,\z_t\}$ satisfies the equality constraint, $\{\x_t,\z_t\}$ need not satisfy $\bA\x_t+ \bB\z_t -\c = \mathbf{0}$. Therefore, we also consider bounds for $R^c$ as defined in \myref{eq:Rc}. A common case we often encounter is when $\bA = \bI, \bB = -\bI, \c = \mathbf{0}$, thus $\hat{\x}_t = \z_t$. Consensus optimization is a typical example of this form~\cite{boyd10,Bertsekas89,nedic10:consensus}. In machine learning, many examples like (group) lasso~\cite{boyd10,Yuan06:grouplasso} can be reformulated in this way.


In this section, we need additional assumptions. In Assumption~\ref{asm:oadm} (a), we specify the dual norm $\|\cdot\|_q$ to be $\ell_2$, i.e., $\|f_t(\x)\|_2\leq G_f$. To guarantee that $\bA\hat{\x}_t + \bB\z_t = \c, \bA \in \R^{m\times n_1}$ is feasible, the equality constraint, in particular, implicitly requires the assumption $m \leq n_1$. On the other hand, to establish a bound for $R_2$, $\bA$ should be full-column rank, i.e., $rank(\bA) = n_1$. Therefore, we need the following assumption in this scenario:

\begin{asm}\label{asm:oadm0}
$\bA$ is a square and full rank matrix, i.e., $\bA$ is invertible. Let $\lambda_{\min}^{\bA}$ be the smallest eigenvalue of $\bA\bA^T$, then $\lambda_{\min}^{\bA} > 0$.
\end{asm}

Assumption~\ref{asm:oadm0} is satisfied in most examples like lasso and consensus optimization. Considering the subgradient of $f_t$ given in~\myref{eq:fg1}, if there always exists a vector $\v_t$ such that $-\bA^T\v_t \in \partial f_t(\x_t)$, Assumption~\ref{asm:oadm0} can be safely removed under the implicit assumption that $\bA\x + \bB\z = \c$ is feasible.

\subsection{General Convex Functions}
The following theorem shows the regret bounds for $R_2$ as well as $R^c$.
%
\begin{thm}\label{thm:oadmm2_rgtbd}
Let $\eta = 0$ in OADM. Let Assumption~\ref{asm:oadm} and~\ref{asm:oadm0} hold. 
For any $\x^*, \z^* $ satisfying $\bA\x^* + \bB\z^* = \c$, setting $\rho = \frac{G_f\sqrt{T}}{D_{\z}\sqrt{\lambda^{\bA}_{\min}\lambda_{\max}^{\bB}}}$, we have
\begin{align}
& R_2(T) \leq \frac{G_fD_{\z}\sqrt{\lambda_{\max}^{\bB}}}{\sqrt{\lambda_{\min}^{\bA}}}\sqrt{T}\label{eq:oadmm2_bd} ~, \\
& R^c(T) \leq \lambda_{\max}^{\bB}D_\z^2 + \frac{2F D_{\z}\sqrt{\lambda^{\bA}_{\min}\lambda_{\max}^{\bB}T}}{G_f}
~. \label{eq:oadmm2_cst1}
\end{align}
\end{thm}
\proof Replacing $f$ by $f_t$ in Lemma \ref{lem:bd1}, we have
\begin{align}\label{eq:oadm2_bd1}
&f_t(\x_{t+1}) + g(\z_{t+1}) - (f(\x^*) + g(\z^*)) \nonumber \\
& \leq \frac{1}{2\rho}(\| \y_{t} \|_2^2 - \| \y_{t+1} \|_2^2)
- \frac{\rho}{2} \| \bA\x_{t+1} + \bB\z_t - \c\|_2^2 + \frac{\rho}{2}  ( \|\bB\z^* - \bB\z_t\|_2^2 -  \|\bB\z^* - \bB\z_{t+1}\|_2^2)~.
\end{align}
Let $f_t'(\hat{\x}_{t}) \in \partial f_t(\hat{\x}_{t})$. Recalling $\bA\hat{\x}_{t} + \bB\z_t = \c$, then
\begin{align}\label{eq:oadm2_fft}
& f_t(\hat{\x}_{t}) - f_t(\x_{t+1}) \leq \langle f_t'(\hat{\x}_{t}), \hat{\x}_{t} - \x_{t+1} \rangle = \langle (\bA^{-1})^T f_t'(\hat{\x}_{t}), \bA\hat{\x}_{t} - \bA\x_{t+1} \rangle \nonumber \\
& = - \langle (\bA^{-1})^T f_t'(\hat{\x}_{t}), \bA\x_{t+1} + \bB\z_t - \c \rangle  \leq \frac{1}{2\lambda^{\bA}_{\min}\rho} \| f_t'(\hat{\x}_{t}) \|_2^2 + \frac{\rho}{2} \| \bA\x_{t+1} + \bB\z_t - \c \|_2^2~.
\end{align}
Adding to \myref{eq:oadm2_bd1} gives
\begin{align}\label{eq:oadm2_fg0}
& f_t(\hat{\x}_{t}) + g(\z_{t+1}) - (f_t(\x^*) + g(\z^*))  \nonumber \\
& \leq \frac{1}{2\lambda^{\bA}_{\min}\rho} \| f_t'(\hat{\x}_{t}) \|_2^2 + \frac{1}{2\rho}(\|\y_t \|_2^2 - \|\y_{t+1} \|_2^2) + \frac{\rho}{2}  ( \|\bB\z^* - \bB\z_t\|_2^2 -  \|\bB\z^* - \bB\z_{t+1}\|_2^2 )~.
\end{align}
Letting the assumptions hold, $R_2(T)$ is bounded as:
\begin{align}\label{eq:R2_logrgt}
R_2(T) & \leq \sum_{t=1}^T \left[ f_t(\hat{\x}_{t}) + g(\z_{t+1})- (f_t(\x^*) + g(\z^*)) \right ] \nonumber \\
& \leq \frac{1}{2\lambda^{\bA}_{\min}\rho} \sum_{t=1}^T\| f_t'(\hat{\x}_{t}) \|_2^2 + \frac{1}{2\rho}(\|\y_1 \|_2^2 - \|\y_{T+1} \|_2^2) + \frac{\rho}{2}  ( \|\bB\z^* - \bB\z_1\|_2^2 -  \|\bB\z^* - \bB\z_{T+1}\|_2^2 ) \nonumber \\
& \leq \frac{G_f^2T}{2\lambda^{\bA}_{\min}\rho} + \frac{ \lambda_{\max}^{\bB}D_{\z}^2\rho }{2}~.
\end{align}
Setting $\rho = \frac{G_f\sqrt{T}}{D_{\z}\sqrt{\lambda^{\bA}_{\min}\lambda_{\max}^{\bB}}}$ yields \myref{eq:oadmm2_bd}.

Now we prove \myref{eq:oadmm2_cst1}. Rearranging the terms in \myref{eq:oadm2_bd1}, we have
\begin{align}\label{eq:cst112}
& \|\bA\x_{t+1} + \bB\z_t - \c\|_2^2 \leq \frac{2F}{\rho} + \frac{1}{\rho^2}(\| \y_{t} \|_2^2 - \| \y_{t+1} \|_2^2) +  ( \|\bB\z^* - \bB\z_t\|_2^2 -  \|\bB\z^* - \bB\z_{t+1}\|_2^2) ~.
\end{align}
Letting the assumptions hold and summing over $t$ from 1 to $T$, we have
\begin{align}
& \sum_{t=1}^T \|\bA\x_{t+1} + \bB\z_t - \c\|_2^2 \nonumber \\
& \leq \frac{2FT}{\rho} + \frac{1}{\rho^2}(\| \y_{1} \|_2^2 - \| \y_{T+1} \|_2^2) +  ( \|\bB\z^* - \bB\z_1\|_2^2 -  \|\bB\z^* - \bB\z_{T+1}\|_2^2)  \nonumber \\
& \leq \frac{2FT}{\rho} + \lambda_{\max}^{\bB}D_\z^2~.
\end{align}
Setting $\rho = \frac{G_f\sqrt{T}}{D_{\z}\sqrt{\lambda^{\bA}_{\min}\lambda_{\max}^{\bB}}}$ and using Lemma \ref{lem:pdr} give \myref{eq:oadmm2_cst1}.
\qed


The following theorem shows that $R^c$ has a constant bound when assuming $\|\y\|_2 \leq D^2$.

\begin{thm}\label{thm:oadmm2_cstbd}
Let the sequences $\{\x_t, \z_t,\y_t\}$ be generated by OADM with $\eta = 0$. Let Assumption~\ref{asm:oadm0} hold. Assuming $\|\y_t\|_2 \leq D^2$ and setting $\rho = \frac{G_f\sqrt{T}}{D_{\z}\sqrt{\lambda^{\bA}_{\min}\lambda_{\max}^{\bB}}}$, we have
\begin{align}
R^c(T) \leq  \frac{2D^2_{\z}\lambda^{\bA}_{\min}\lambda_{\max}^{\bB}} {G_f^2} (D^2 + \frac{G_f^2}{\lambda_{\min}^{\bA}})~. \label{eq:oadmm2_cst2}
\end{align}
\end{thm}
\proof
Let $f$ be $f_t$ in \myref{eq:fg1}. Define
\begin{align}
f_t'(\x_{t+1}) = -(\bA^T\y_t + \rho\bA^T(\bA\x_{t+1} + \bB\z_t - \c))~.
\end{align}
Multiplying both sides by $(\bA^T)^{-1}$ gives
\begin{align}
(\bA^T)^{-1}f_t'(\x_{t+1}) = -(\y_t + \rho(\bA\x_{t+1} + \bB\z_t - \c))~.
\end{align}
Rearranging the terms, we have
\begin{align}\label{eq:cst211}
\|\bA\x_{t+1} + \bB\z_t - \c \|_2^2 & = \frac{1}{\rho^2} \| \y_t + (\bA^T)^{-1}f_t'(\x_{t+1})\|_2^2 \nonumber \\
& \leq \frac{2}{\rho^2} (\| \y_t\|_2^2 + \|(\bA^T)^{-1}f_t'(\x_{t+1})\|_2^2) \nonumber \\
& \leq \frac{2}{\rho^2} (D^2 + \frac{G_f^2}{\lambda_{\min}^{\bA}})~.
\end{align}
Summing over $t$ from 1 to T and setting $\rho = \frac{G_f\sqrt{T}}{D_{\z}\sqrt{\lambda^{\bA}_{\min}\lambda_{\max}^{\bB}}}$, we have \myref{eq:oadmm2_cst2} according to Lemma 2.
\qed

Without requiring an additional Bregman divergence, $R_2$ achieves the same $\sqrt{T}$ bound as $R_1$. While $R_1$ depends on $\x_t$ which may not stay in the feasible set, $R_2$ is defined on $\hat{\x}_t$ which always satisfies the equality constraint. The corresponding algorithm requires finding $\hat{\x}_t$ in each iteration such that $\bA \hat{\x}_t = \c - \bB \z_t$, which involves solving a linear system. The algorithm will be efficient in some settings, e.g., consensus optimization where $\bA = \bI$.

\subsection{Strongly Convex Functions}

If $g(\z)$ is a $\beta_2$-strongly convex function given in \myref{eq:gsc}, we show that $R_2$ and $R^c$ have logarithmic bounds.

\begin{thm} Let $\eta = 0$ in OADM. Assume that $g(\z)$ is $\beta_2$-strongly convex and Assumption~\ref{asm:oadm} and~\ref{asm:oadm0} hold. Setting $\rho_t = \beta_2 t/\lambda_{\max}^{\bB}$, we have
\begin{align}
& R_2(T) \leq \frac{G_f^2\lambda_{\max}^{\bB}}{2\lambda^{\bA}_{\min}\beta_2} (\log(T+1))  + \beta_2 D_{\z}^2 ~,\label{eq:logrgt2} \\
& R^c(T) \leq \lambda_{\max}^{\bB}D_\z^2 + \frac{2F \lambda_{\max}^{\bB}}{\beta_2}\log(T+1) ~.\label{eq:logrgt2_cst}
\end{align}
\end{thm}
\proof Assuming $g(\z)$ is strongly convex~\myref{eq:gsc},  we can show the regret bound by replacing $\rho$ by $\rho_{t+1}$ and subtracting the strongly convex term $\frac{\beta_2}{2}\|\z^* - \z_{t+1}\|_2^2$ in~\myref{eq:oadm2_fg0}, i.e., 
\begin{align}
& f_t(\hat{\x}_{t}) + g(\z_{t+1}) - (f_t(\x^*) + g(\z^*))  \leq \frac{1}{2\lambda^{\bA}_{\min}\rho_{t+1}} \| f_t'(\hat{\x}_{t}) \|_2^2 + \frac{1}{2\rho_{t+1}}(\|\y_t \|_2^2 - \|\y_{t+1} \|_2^2)  \nonumber \\
&+ \frac{\rho_{t+1}}{2}  ( \|\bB\z^* - \bB\z_t\|_2^2 -  \|\bB\z^* - \bB\z_{t+1}\|_2^2 )  - \frac{\beta_2}{2}\|\z^* - \z_{t+1}\|_2^2~.
\end{align}
Summing over $t$ from 1 to $T$, we have
\begin{align}
R_2(T) & \leq \frac{G_f^2}{2\lambda^{\bA}_{\min}} \sum_{t=1}^T \frac{1}{\rho_{t+1}} + \sum_{t=1}^T \frac{1}{2\rho_{t+1}}(\|\y_t \|_2^2 - \|\y_{t+1} \|_2^2) \nonumber \\
& + \sum_{t=1}^T\left[ \frac{\rho_{t+1}}{2}  ( \|\bB\z^* - \bB\z_t\|_2^2 -  \|\bB\z^* - \bB\z_{t+1}\|_2^2 ) - \frac{\beta_2}{2}\|\z^* - \z_{t+1}\|_2^2 \right] ~.
\end{align}
Using \myref{eq:yrho}, \myref{eq:logrgt_z} and setting $\rho_t = \beta_2 t/\lambda_{\max}^{\bB}$, we get \myref{eq:logrgt2} by applying $\sum_{t=1}^T \frac{1}{t+1} \leq \log(T+1)$.

Now we prove \myref{eq:logrgt2_cst}. Replacing $\rho$ by $\rho_{t+1}$ in \myref{eq:cst112}, we have
\begin{align}
& \|\bA\x_{t+1} + \bB\z_t - \c\|_2^2 \leq \frac{2F}{\rho_{t+1}} + \frac{1}{\rho_{t+1}^2}(\| \y_{t} \|_2^2 - \| \y_{t+1} \|_2^2) +  ( \|\bB\z^* - \bB\z_t\|_2^2 -  \|\bB\z^* - \bB\z_{t+1}\|_2^2) ~.
\end{align}
Letting the assumptions hold and summing over $t$ from 0 to $T$, we have
\begin{align}
& \sum_{t=1}^T \|\bA\x_{t+1} + \bB\z_t - \c\|_2^2  \nonumber \\
& \leq 2F\sum_{t=1}^T\frac{1}{\rho_{t+1}} + \sum_{t=1}^T\frac{1}{\rho_{t+1}^2}(\| \y_{t} \|_2^2 - \| \y_{t+1} \|_2^2) +  ( \|\bB\z^* - \bB\z_1\|_2^2 -  \|\bB\z^* - \bB\z_{T+1}\|_2^2 )  \nonumber \\
& \leq 2F\sum_{t=1}^T\frac{1}{\rho_{t+1}} + \lambda_{\max}^{\bB}D_\z^2~.
\end{align}
We use \myref{eq:yrho} in the last inequality. Setting $\rho_t = \beta_2 t/\lambda_{\max}^{\bB}$ and using Lemma \ref{lem:pdr} give \myref{eq:logrgt2_cst}.
\qed

Similar as the case of general convex functions, the logarithmic regret bound for constraint violation can also be reduced to a constant bound, as shown in the following theorem.
\begin{thm}\label{thm:lgobd2}
Let $\eta = 0$ in OADM. Assume that $g(\z)$ is $\beta_2$-strongly convex and Assumption~\ref{asm:oadm0} hold.  Assuming $\| \y_t \|_2 \leq D$ and setting $\rho_t = \beta_2 t/\lambda_{\max}^{\bB}$, we have
\begin{align}
R^c(T) \leq \frac{\pi{\lambda_{\max}^{\bB}}^2}{3\beta_2^2} \left(D^2 + \frac{G_f^2}{\lambda_{\min}^{\bA}} \right)\label{eq:logcst}
\end{align}
\end{thm}
\proof
Setting $\rho_t = \beta_2 t/\lambda_{\max}^{\bB}$ in \myref{eq:cst211}, summing over $t$ from 1 to $T$ and using $\sum_{t=1}^T \frac{1}{t^2} \leq \frac{\pi}{6}$ complete the proof.
\qed 

Theorem \ref{thm:lgobd2} shows that OADM can achieve the logarithmic regret bound without requiring $f_t$ to be strongly convex, which is in line with other online learning algorithms for composite objectives.
%
%

\section{Further Discussions}
In this section, we discuss several variants of the $\x$ update in OADM which can lead to efficient updates and show the stochastic convergence rates. The connection to the related work is presented. We also show that OADM can serve as projection-free online learning.

\subsection{Inexact ADMM Updates ($\eta > 0$)}\label{sec:inexactADMM}

In OADM ($\eta > 0$), since the $\x$ update~\myref{eq:oadmm1_x} involves the function $f_t$,  the quadratic penalty term and a Bregman divergence,  it may be computationally expensive to solve it exactly. We consider several  variants which solve the $\x$ update inexactly through the linearization of some terms. The inexact updates can be efficient, and include mirror descent  algorithm (MDA) and composite objective mirror descent (COMID) as special cases. 

\textbf{Case 1: Linearization of the quadratic penalty term} The linearization of the quadratic penalty term in~\myref{eq:oadmm1_x} can be done by removing $\| \bA\x\|_2^2$ as follows:
\begin{align*}
\| \bA\x + \bB\z_t - \c\|_2^2 - \|\bA(\x - \x_t)\|_2^2 = 2\langle \bA\x_t + \bB\z_t - \c, \bA\x\rangle +  \|\bB\z_t - \c\|_2^2 - \|\bA\x_t\|_2^2~.
\end{align*}
Let $B_{\phi}(\x,\x_t) = B_{\varphi}(\x,\x_t) - \frac{\rho}{2\eta}\|\bA(\x - \x_t)\|_2^2$ in~\myref{eq:oadmm1_x}, where $B_{\varphi}$ is a Bregman divergence and the quadratic term is used to linearize the quadratic penalty term. Removing constant terms, \myref{eq:oadmm1_x} becomes
\begin{align}\label{eq:snr2case2}
\x_{t+1} = \argmin_\x f_t(\x) + \langle \y_t + \rho (\bA\x_t + \bB\z_t - \c), \bA\x\rangle + \eta B_{\varphi}(\x, \x_t)~.
\end{align}
This case mainly solves the problem caused by $\bA$, e.g., $\bA\x$ makes $\x$ nonseparable. Several problems have been benefited from the linearization of quadratic term~\cite{deng12:admm}, e.g., $f$ is $\ell_1$ loss function~\cite{Hastie09} and projection onto the unit simplex or $\ell_1$ ball~\cite{Duchi08}.

Since $B_{\phi}(\x,\x_t) \geq \frac{\alpha}{2}\|\x - \x_t\|_2^2$ is required for the analysis in Section~\ref{sec:oadmm1},  $B_{\varphi}$ should be chosen to satisfy that condition. Note  
\begin{align}
B_{\phi}(\x,\x_t)  = B_{\varphi}(\x,\x_t) - \frac{\rho}{2\eta}\|\bA(\x - \x_t)\|_2^2 \geq B_{\varphi}(\x,\x_t)  - \frac{\rho\lambda_{\max}^{\bA}}{2\eta} \|\x - \x_t\|_2^2~.
\end{align}
Therefore, as long as $ B_{\varphi}(\x,\x_t)  \geq \frac{\rho\lambda_{\max}^{\bA}/\eta + \alpha}{2} \|\x - \x_t\|_2^2$, the assumption~\ref{asm:oadm}(b) holds, meaning Theorem~\ref{thm:oadmm1_rgtbd} and~\ref{thm:oadm_logrgtbd1} hold for Case 1.

\textbf{Case 2: Linearization of function $f_t$}  This case is particularly useful when the difficulty of solving~\myref{eq:oadmm1_x} is caused by $f_t(\x)$, e.g., when $f_t$ is a logistic loss function. Linearizing the function $f_t$ at $\x_t$ in~\myref{eq:oadmm1_x}, we have 
\begin{align}\label{eq:snr2case2}
\x_{t+1} = \argmin_\x \langle f'_t(\x_t), \x - \x_t\rangle + \frac{\rho}{2} \| \bA\x + \bB\z_t - \c \|_2^2 + \eta B_{\phi}(\x, \x_t)~.
\end{align}
The updated is called inexact ADMM update if $\phi$ is a quadratic function~\cite{boyd10}.  In the Appendix~\ref{sec:app1},  we show Theorem~\ref{thm:oadmm1_rgtbd} and~\ref{thm:oadm_logrgtbd1} continue to hold in this case. 

\textbf{Case 3: Mirror Descent}  In this case, we linearize both the function and the quadratic term, which can be done by choosing $B_{\phi}(\x,\x_t) = B_{\varphi}(\x,\x_t)  - \frac{\rho}{2\eta}\|\bA(\x - \x_t)\|_2^2$ in Case 2. Combining the results in Case 1 and 2,~\myref{eq:oadmm1_x} becomes the following MDA-type update:
\begin{align}\label{eq:snr2case3}
\x_{t+1} &= \argmin_\x \langle F_t(\x_t), \x \rangle + \eta B_{\varphi}(\x, \x_t)~,
\end{align}
where $F_t(\x_t) = f'_t(\x_t) + \bA^T\{\y_t + \rho(\bA\x_t + \bB\z_t - \c)\}$, which is the gradient of the objective in ~\myref{eq:oadmm1_x}. Assuming $ B_{\varphi}(\x,\x_t)  \geq \frac{\rho\lambda_{\max}^{\bA}/\eta + \alpha}{2} \|\x - \x_t\|_2^2$ in Case 2,  the regret bounds in Theorem~\ref{thm:oadmm1_rgtbd} and~\ref{thm:oadm_logrgtbd1}  still holds in Case 3.

\textbf{Case 4: COMID} Assume $f_t$ is a composite objective consisting of smooth and nonsmooth part, i.e., $f_t(\x) = f_t^{S}(\x) + f_t^{N}(\x)$, where $f_t^S$ is the smooth part and $f_t^{N}$ is the nonsmooth part.  Let $B_{\phi}(\x,\x_t) = B_{\varphi}(\x,\x_t) - \frac{\rho}{2\eta}\|\bA(\x - \x_t)\|_2^2$, which is used to linearize the quadratic penalty term.  Linearizing the smooth function $f_t^S$,~\myref{eq:oadmm1_x} becomes the following COMID-type update:
\begin{align}\label{eq:snr2case4}
\x_{t+1} &= \argmin_\x  f_t^{N}(\x)+ \langle F_t^S(\x_t), \x \rangle + \eta B_{\varphi}(\x, \x_t)~,
\end{align}
where $F_t^{S}(\x_t) = \nabla f_t^{S}(\x_t) + \bA^T\{\y_t + \rho(\bA\x_t + \bB\z_t - \c)\}$.  Applying the analysis in Case 2 on the smooth part, we can get the regret bounds in Theorem~\ref{thm:oadmm1_rgtbd} and~\ref{thm:oadm_logrgtbd1}.

\subsection{Stochastic Convergence Rates}
In this section, we present the convergence rates for ADMM in the Case 2-4 in Section 6.1 in the stochastic setting, which solves the following stochastic learning problem:
\begin{align}
\min_{\x \in \mathcal{X},\z\in\mathcal{Z}} \E_{\xi}[f(\x,\xi)] + g(\z)\quad \text{s.t.} \quad \bA\x + \bB\z = \c
\end{align}
$f'(\x_t,\xi_t)$ is an unbiased estimate of $f'(\x_t)$ and $f(\x)  = \E f(\x,\xi)$. Correspondingly,  the $\x$-update in~\myref{eq:snr2case2}-\myref{eq:snr2case3} uses $f'(\x_t,\xi_t)$ to substitute $f'_t(\x_t)$ and $\nabla f^N(\x_t,\xi_t)$ to substitute $\nabla f^N_t(\x_t)$ in~\myref{eq:snr2case4}. 
The regret bounds for Case 2-4 in Section 6.1 can be converted to convergence rates in the stochastic setting based on known online-stochastic conversion~\cite{Cesa04:stochastic,duchi09,xiao10}. More specifically, the stochastic convergence rates in expectation can be obtained by simply dividing regret bounds by $T$. Using martingale concentration results~\cite{Cesa04:stochastic,duchi09,xiao10}, the high probability bounds can also be obtained by applying the Azuma-Hoeffding inequality~\cite{azuma}. 

\begin{cor}\label{cor:sadm1}
Let the sequences $\{\x_t, \z_t,\y_t\}$ be generated by stochastic ADM and Assumption~\ref{asm:oadm} hold. Let $\bar{\x}_T = \frac{1}{T}\sum_{t=1}^T\x_t$ and $\bar{\z}_T = \frac{1}{T}\sum_{t=1}^T\z_t$. For any $\x^*,\z^*$ satisfying $\bA\x^* + \bB\z^* = \c$, setting $\eta = \frac{G_f\sqrt{T}}{D_\x \sqrt{2\alpha}}$ and $\rho = \sqrt{T}$,  we have

\noindent (a) Stochastic convergence rates in expectation
\begin{align}
&\E\left[f(\bar{\x}_T)+g(\bar{\z}_T)] - (f(\x^*) + g(\z^*)) \right]
\leq \frac{\lambda_{\max}^{\bB}D_{\z}^2}{2\sqrt{T}} + \frac{\sqrt{2}G_fD_\x}{\sqrt{\alpha}\sqrt{T}}  ~,\label{eq:sadmm_eobj} \\
&\E\left[\| \bA\bar{\x}_{T} + \bB\bar{\z}_{T} + \c\|_2^2 \right] \leq \frac{\lambda_{\max}^{\bB}D_\z^2}{T} + \frac{2\sqrt{2}D_{\x}G_f}{\sqrt{\alpha}T} + \frac{2F}{\sqrt{T}} \label{eq:sadmm_ecst}~.
\end{align}
(b) High probability bounds for stochastic convergence rates
\begin{align}
&P\big(f(\bar{\x}_T)+g(\bar{\z}_T) - (f(\x^*) + g(\z^*)) \geq \frac{\lambda_{\max}^{\bB}D_{\z}^2}{2\sqrt{T}} + \frac{\sqrt{2}G_fD_\x}{\sqrt{\alpha}\sqrt{T}} + \varepsilon \big) \leq  \exp\left(-\frac{T\alpha\varepsilon^2}{16D^2_\x G_f^2}\right)\label{eq:sadm2_bd} ~, \\
&P\big(\|\bA\bar{\x}_{T} + \bB\bar{\z}_T - \c\|_2^2  \geq \frac{2F}{\sqrt{T}} + \frac{\lambda_{\max}^{\bB}D_\z^2}{T} +\frac{2\sqrt{2}D_{\x}G_f}{\sqrt{\alpha}T} + \varepsilon \big )\leq  \exp\left(-\frac{T\alpha\varepsilon^2}{16D^2_\x G_f^2}\right)~.\label{eq:sadm2_cst}
\end{align}
\end{cor}
The proof is presented in Appendix~\ref{sec:app2}. Compared to the stochastic convergence rates for COMID~\cite{Duchi10_comid}, the stochastic convergence rates for the objective of ADM has an extra term $\frac{\lambda_{\max}^{\bB}D_{\z}^2}{2\sqrt{T}}$  which bounds the splitting variable $\z$.  For strongly convex functions, we have $O(\frac{\log T}{T})$ stochastic convergence rates by applying the online-stochastic conversion~\cite{Cesa04:stochastic,duchi09,xiao10} on Theorem~\ref{thm:oadm_logrgtbd1}.

\begin{rmk}
 We note that~\cite{ou13:sadmm} has recently established the stochastic convergence rates for stochastic ADM based on our VI analysis (see Section 2.3), which has the following form in our notation:
 \begin{align}\label{eq:sadmmou}
 \E\left[f(\bar{\x}_T)+g(\bar{\z}_T)\! -\! (f(\x^*) + g(\z^*)) \!+\! D\| \bA\bar{\x}_{T} + \bB\bar{\z}_{T} + \c\|_2^2\right] \!\leq\! \frac{\lambda_{\max}^{\bB}D_{\z}^2\rho}{2T} \!+\! \frac{\sqrt{2}G_fD_\x}{\sqrt{T}} \!+\! \frac{D^2}{2\rho T}~,
 \end{align}
 where $\|\y_t\|_2 \leq D$ (see Assumption~\ref{asm:admvi}).  The bound in~\myref{eq:sadmmou} depends on $D^2$, which usually is large (see Eq.~\myref{eq:yT}) and thus worse than our results which do not rely on $D^2$.   As a matter of fact, we can show the term $D^2$ can be safely removed (setting $\alpha =1$ in~\myref{eq:sadmm_objcst} in Appendix~\ref{sec:app2}), i.e.,
  \begin{align}\label{eq:sadmmwang}
\!\!\!\!\!\!\!\!\E\left[f(\bar{\x}_T)+g(\bar{\z}_T) \!-\! (f(\x^*) + g(\z^*)) \!+\! \frac{\rho}{2}\| \bA\bar{\x}_{T} + \bB\bar{\z}_{T} + \c\|_2^2\right] \!\leq\! \frac{\lambda_{\max}^{\bB}D_{\z}^2\rho}{2T} \!+\! \frac{\sqrt{2}G_fD_\x}{\sqrt{T}}~.
   \end{align}  
However, since $\x_t,\z_t$ are not feasible, $f(\bar{\x}_T)+g(\bar{\z}_T) - (f(\x^*) + g(\z^*))$ may be negative. As a result,~\myref{eq:sadmmou} or~\myref{eq:sadmmwang} may not imply an $O(1/T)$ convergence rate for the equality constraint, in constrast to~\myref{eq:sadmm_ecst} in Corollary~\ref{cor:sadm1}. Furthermore, if assuming $\|\y_t\|_2 \leq D$ , the residual of equality constraint has an $O(1/T)$ convergence rate by dividing by $T$ on both sides of~\myref{eq:oadmm1_cst1} in Theorem~\ref{thm:cstbd} and using the Jensen's inequality.
  \end{rmk}

\subsection{Connections to Related Work ($\eta = 0$)}
Assume $\eta = 0, \bA = \bI, \bB = -\bI, \c = \mathbf{0}$, thus $\x = \z$. Hence, the online optimization problem has the form which is the same as the ones considered in the development of FOBOS~\cite{duchi09} and RDA~\cite{xiao10}.
The three steps of OADM ($\eta = 0$) reduce to
\begin{align}
\x_{t+1} & = \underset{\x}{\argmin} \{f_t(\x) + \langle \y_t, \x - \z_t \rangle + \frac{\rho}{2} \| \x - \z_t \|_2^2\}~,  \label{eq:oadmm_x1} \\
\z_{t+1} & = \underset{\z}{\argmin} \{g(\z) + \langle \y_t, \x_{t+1} - \z \rangle + \frac{\rho}{2} \| \x_{t+1} - \z\|_2^2\}~, \label{eq:oadmm_z1}\\
\y_{t+1} & = \y_t + \rho(\x_{t+1} - \z_{t+1})~. \label{eq:oadmm_y1}
\end{align}
Let $f_t'(\x_{t+1}) \in \partial f_t(\x), g'(\z_{t+1}) \in \partial g(\z)$. The first order optimality conditions for \myref{eq:oadmm_x1} and \myref{eq:oadmm_z1} give
\begin{align*}
& f_t'(\x_{t+1}) + \y_t + \rho(\x_{t+1} - \z_t) = 0 ~,\\
&  g'(\z_{t+1}) - \y_t - \rho(\x_{t+1}-\z_{t+1}) = 0~.
\end{align*}
Adding them together yields
\begin{align}
\z_{t+1} = \z_t - \frac{1}{\rho} (f_t'(\x_{t+1}) + g'(\z_{t+1}))~.
\end{align}
OADM can be considered as taking the implicit subgradient of $f_t$ and $g$ at the yet to be determined $\x_{t+1}$ and $\z_{t+1}$. FOBOS has the following update \cite{duchi09}:
\begin{align*}
\z_{t+1} = \z_t - \frac{1}{\rho} (f_t'(\z_{t}) + g'(\z_{t+1}))~.
\end{align*}
FOBOS takes the explicit subgradient of $f_t$ at current $\z_t$.
In fact, FOBOS can be considered as a variant of OADM, which linearizes the objective of \myref{eq:oadmm_x1} at $\z_t$ :
\begin{align*}
\x_{t+1} &= \underset{\x}{\argmin} \langle f_t'(\z_t) + \y_t, \x - \z_t \rangle + \frac{\rho}{2} \| \x - \z_t \|_2^2~.
\end{align*}
It has a closed-form solution, i.e.,  $\x_{t+1} = \z_t - \frac{1}{\rho}(f_t'(\z_t)+ \y_t)$. Denote $\z_{t+\frac{1}{2}} = \x_{t+1} + \frac{1}{\rho}\y_t$, then
\begin{align}\label{eq:admm_fobos1}
& \z_{t+\frac{1}{2}} = \z_t - \frac{1}{\rho}f_t'(\z_t)~.
\end{align}
\myref{eq:oadmm_z1} is equivalent to the following form:
\begin{equation} \label{eq:gz1}
\z_{t+1} = \argmin_{\z} g(\z) + \frac{\rho}{2} \| \z - \z_{t+\frac{1}{2}}\|_2^2~.
\end{equation}
\myref{eq:admm_fobos1} and~\myref{eq:gz1} form the updates of FOBOS~\cite{duchi09}.
Furthermore, if $g(\z)$ is an indicator function of a convex set $\Omega$, substituting \myref{eq:admm_fobos1} into \myref{eq:gz1}, we have
\begin{align*}
\z_{t+1} &= \argmin_{\z\in\Omega} \frac{\rho}{2} \| \z_t - \frac{1}{\tau}f_t'(\z_t) - \z \|_2^2 = \mathcal{P}_{\z\in\Omega}\left[ \z_t - \frac{1}{\tau}f_t'(\z_t) \right]~,
\end{align*}
and we recover projected gradient descent~\cite{haak07}.  

\subsection{Projection-free Online Learning}\label{sec:linie}
For an online constrained optimization problem, the state-of-the-art methods like OGD, FOBOS and RDA require a full projection onto the constraint set at each round. In many cases, e.g., an intersection of simple constraints, the full projection can be done by alternating projecting onto simple constraints cyclically~\cite{ceze98}. In OADM, we can decompose functions and constraints into simpler subproblems by introducing appropriate splitting variables. If the subproblem for each splitting variable is simple enough to yield efficient projection, the full projection onto the whole constraint set can be done by projections onto simple constraints at each round along with the long term equality constraints. Therefore, OADM and its variants can avoid the full projection at each round.
Consider the full projection onto $\mathcal{X}\times \mathcal{Z}$, which in general requires alternating projection onto $\cX$ and $\cZ$ at each round in OGD, FOBOS and RDA. In OADM, by introducing equality
constraint $\x = \z$, the constraint set is split into two parts and $\x \in \mathcal{X}$ and $\z \in \mathcal{Z}$. At each round, the primal updates in OADM and its variants project $\x,\z$ onto $\mathcal{X},\mathcal{Z}$ separately. In the long run, the equality
constraint will be satisfied in expectation, thus $\x$ is a feasible solution. Hence, OADM can be considered as a projection-free online learning algorithm.

In~\cite{hazan12:free}, the Frank-Wolfe algorithm is used as a projection-free online learning algorithm, which solves a linear optimization at each round and has $O(T^{3/4})$ regret bound. It assumes linear optimization can be done efficiently in the constraint set. Realizing that
solving a linear optimization still requires an inner loop algorithm, the authors pose an open problem whether the optimal
regret bound can be achieved by performing one iteration of linear-optimization. 

We now show how OADM does projection-free online learning with linear constraints, which includes linear
programming and quadratic programming as special cases. Formally, we consider the problem
\begin{align}\label{eq:lc}
\min_{\x} \sum_{t=1}^T f_t(\x) \quad \text{s.t.} \quad \bA\x = \a, \bB\x \leq \b~.
\end{align}
In the setting of OADM, we first introduce an auxiliary variable $\z = \bB\x$ to separate inequality constraint from equality constraint. Then \myref{eq:lc} can be rewritten as:
\begin{align}\label{eq:lc_oadm}
\min_{\x,\z} \sum_{t=1}^T f_t(\x) + g(\z) \quad \text{s.t.} \quad \bA\x = \a, \bB\x = \z~,
\end{align}
where $g(\z)$ is the indicator function of box constraint $\z \leq \b$.
The augmented Lagrangian for~\myref{eq:lc_oadm} is as follows:
\begin{align}
L_{\rho}(\x,\z,\u,\v) &= f_t(\x) + g(\z) + \langle \u, \bA\x - \a\rangle + \langle \v, \bB\x - \z \rangle \nonumber \\
&+ \frac{\rho_\u}{2} \| \bA\x - \a \|_2^2 + \frac{\rho_\v}{2} \| \bB\x - \z \|_2^2~,
\end{align}
where $\u,\v$ are dual variables and the penalty parameters $\rho_\u, \rho_\v > 0$. Let the Bregman divergence in the $\x$ update in~\myref{eq:oadmm1_x} be the quadratic function. We have the following OADM updates for \myref{eq:lc_oadm}:

\begin{align}
& \x_{t+1} = \underset{\x}{\argmin} \left \{ f_t(\x) + \langle \u_t, \bA\x - \a\rangle + \langle \v_t, \bB\x - \z_t \rangle + \frac{\rho_\u}{2} \| \bA\x - \a \|_2^2 \right. \nonumber \\
& \quad \quad \quad \quad \quad \quad  \left . + \frac{\rho_\v}{2} \|\bB\x - \z_t \|_2^2 + \frac{\eta}{2} \| \x - \x_t \|_2^2 \right \}~, \\
& \z_{t+1} = \underset{\z \leq \b}{\argmin}~\left \{ \langle \v_t, \bB\x_{t+1} - \z \rangle  + \frac{\rho_\v}{2} \|\bB\x_{t+1} - \z \|_2^2 \right\}~, \\
& \u_{t+1} = \u_t + \rho_\u( \bA\x_{t+1} - \a)~, \\
& \v_{t+1} = \v_t + \rho_\v(\bB\x_{t+1} - \z_{t+1})~,
\end{align}
where $\eta \geq 0$.
The $\x$-update has a closed-form solution when $f_t$ is a linear or quadratic functions, or the $\ell_1$ norm. If the $\x$-update does not have a closed-form solution, we can linearize $f_t$ at $\x_t$ as in Section \ref{sec:inexactADMM}, which leads to a closed-form solution.  Further, the $z$-update has a closed-form solution of the following form:
\begin{align}
\z_{t+1} = \min\{ B\x_{t+1} + \y_t/\rho,\b \}~.
\end{align}
Thus, OADM gives a projection-free online algorithm for optimization problems under linear constraints, e.g., linear and quadratic programming. In contrast, state-of-the-art online learning algorithms require the projection onto the constraints at each round, which amounts to solving a linear or quadratic program~\cite{hazan12:free}.


\section{Experimental Results}
In this section, we use OADM to solve generalized lasso problems~\cite{boyd10}, including lasso~\cite{Tibshirani96} and total variation (TV) problem~\cite{Rudin92:tv}. We present simulation results to show the convergence of the objective as well as constraints in OADM. We also compare it with batch ADM and two other online learning algorithms: FOBOS~\cite{duchi09} and regularized dual averaging (RDA)~\cite{xiao10} in selecting sparse dimension in lasso and recovering data in total variation.

\subsection{Generalized Lasso}
The generalized lasso problem is formulated as follows:
\begin{align}\label{eq:glasso}
\min_{\x} \frac{1}{N}\sum_{t=1}^N\| \a_t\x - b_t \|_2^2 + \lambda | \bD\x |_1~,
\end{align}
where $\a_t\in\R^{1\times n}, \x \in \R^{n\times 1}, \bD \in \R^{m\times n}$ and $b_t$ is a scalar. If $\bD = \bI$, \myref{eq:glasso} yields the lasso. If $\bD$ is an upper bidiagonal matrix with  diagonal $1$ and off-diagonal $-1$,
\myref{eq:glasso} becomes the problem of total variation. The ADM form of \myref{eq:glasso} is:
\begin{align}
\min_{\bD\x = \z} \frac{1}{N}\sum_{t=1}^N\| \a_t\x - b_t \|_2^2 + \lambda | \z |_1~,
\end{align}
where $\z \in \R^{m\times 1}$. The augmented Lagrangian at round $t$ is
\begin{align*}
L_{\rho} = \| \a_t\x - b_t \|_2^2 + \lambda | \z |_1 + \langle \y, \bD\x - \z \rangle + \frac{\rho}{2} \|\bD\x - \z\|_2^2~.
\end{align*}
The three updates of OADM yield the following closed-form updates:
\begin{align}
& \x_{t+1} = (\a_t^T\a_t + \rho \bD^T\bD + \eta)^{-1}\v_t ~, \label{eq:lasso_x}\\
& \z_{t+1}  = S_{\lambda/\rho} (\bD\x_{t+1} + \u_t) ~,  \label{eq:lasso_z} \\
& \u_{t+1} = \u_t + \bD\x_{t+1} - \z_{t+1} ~, \label{eq:lasso_y}
\end{align}
where $\u = \y/\rho$, $\v_t = \a_{t}^Tb_t + \rho \bD^T(\z_t - \u_t) + \eta\x_t$, and $S_{\lambda/\rho}$ denotes the soft thresholding operator or a shrinkage operator defined as
\begin{align}
S_{\lambda/\rho}(k) = \left \{
\begin{array}{cc}
k - \lambda/\rho, & k > \lambda/\rho \\
0, & |\x| \leq \lambda/\rho \\
k + \lambda/\rho, & k < - \lambda/\rho
\end{array} \right. ~,
\end{align}
which is a simple element-wise operation.

For lasso, the $\x$-update is
\begin{align*}
\x_{t+1} = (\v_t - (\eta+\rho+\a_t\a_t^T)^{-1}\a_t^T(\a_t\v_t))/(\eta+\rho)~,
\end{align*}
where the inverse term is a scalar. The multiplication terms take $O(n)$ flops~\cite{GOLO96_matrix}. Thus, the $\x$-update can be done in $O(n)$ flops.

For total variation, we set $\eta = 0$ so that
\begin{align*}
\x_{t+1} = (\bQ\v_t - (\rho + \a_t\bQ\a_t^T)^{-1}\bQ\a_t^T(\a_t\bQ\v_t))/\rho~,
\end{align*}
where $\bQ = (\bD^T\bD)^{-1}$. Since $\bD$ is a bidiagonal matrix, $\bQ\v_t$ and $\bQ\a_t$ can be done in $O(n)$ flops~\cite{GOLO96_matrix,boyd10}. The inverse term is scalar and other multiplication terms cost $O(n)$ flops. Overall, the $\x$-update can be carried out in $O(n)$ flops.

In both cases, the three updates \myref{eq:lasso_x}-\myref{eq:lasso_y} can be done in $O(n)$ flops. In contrast, in batch ADM, the complexity of $\x$-update could be as high as $O(n^3)$ or $O(n^2)$ by caching factorizations \cite{boyd10}.

FOBOS and RDA cannot directly solve the TV term.  We first reformulate the total variation in the lasso form such that
\begin{align}
\min_{\y} \frac{1}{N}\sum_{t=1}^N\| \a_t\bD^{-1}\y - \b \|_2^2 + \lambda | \y |_1~,
\end{align}
where $\y = \bD\x$. FOBOS and RDA can solve the above lasso problem and get $\y$. $\x$ can be recovered by using $\x = \bD^{-1}\y$.

\subsection{Simulation}
Our experiments mainly follow the lasso and total variation examples in~\cite{boyd10},\footnote{\url{http://www.stanford.edu/~boyd/papers/admm/}} although we modified the code to accommodate our setup. We first randomly generated $\bA$ with $N$ examples of dimensionality $n$. $\bA$ is then normalized along the columns. Then, a true $\x_0$ is randomly generated with certain sparsity pattern for lasso and TV. For lasso, we set the number of nonzeros (NNZs) $k$ in $\x_0$ as $100$, i.e., $k = 100$. For TV, we first set $\x_0$ to be a vector of ones, then randomly select some blocks of random size in $\x_0$ and reset their value to a random value from $[1,10]$. $\b$ is calculated by adding Gaussian noise to $\bA\x_0/N$. In all experiments, $N = 100$, which facilitates the matrix inverse in ADM. For lasso, we try different combination of parameters from $n = [1000,5000]$, $\rho = [0.1,1,10]$ and $q = [0.1,0.5]$ for $\lambda = q\times |\bA^Tb/N|_{\infty}$. All experiments are implemented in Matlab.

\begin{figure}[!t]
\subfigure[Sparsity.]{ \label{fig:sparse_best}
\includegraphics[width = 49mm,height = 35mm]{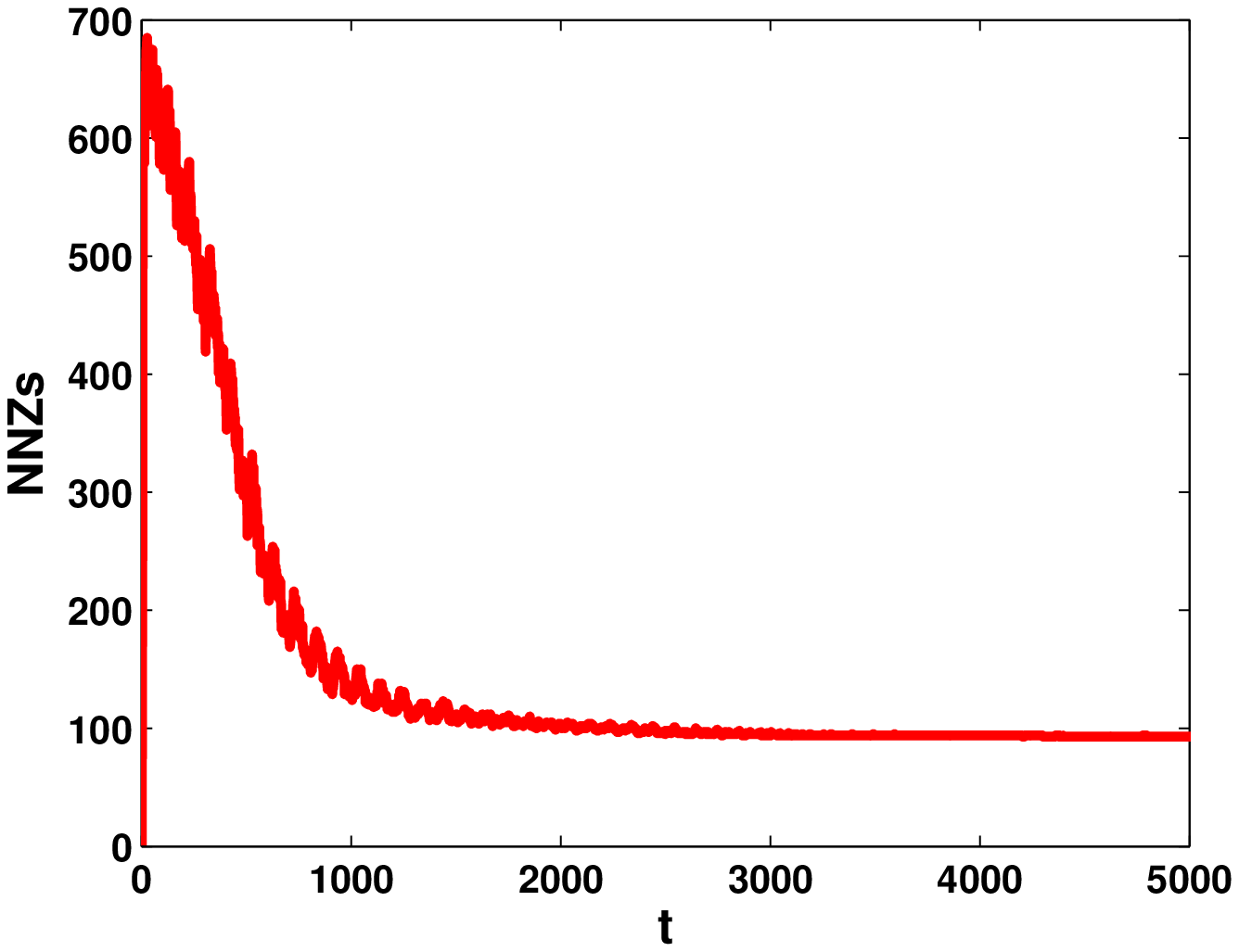}}
\subfigure[Objective.]{ \label{fig:obj_best}
\includegraphics[width= 49mm,height = 35mm]{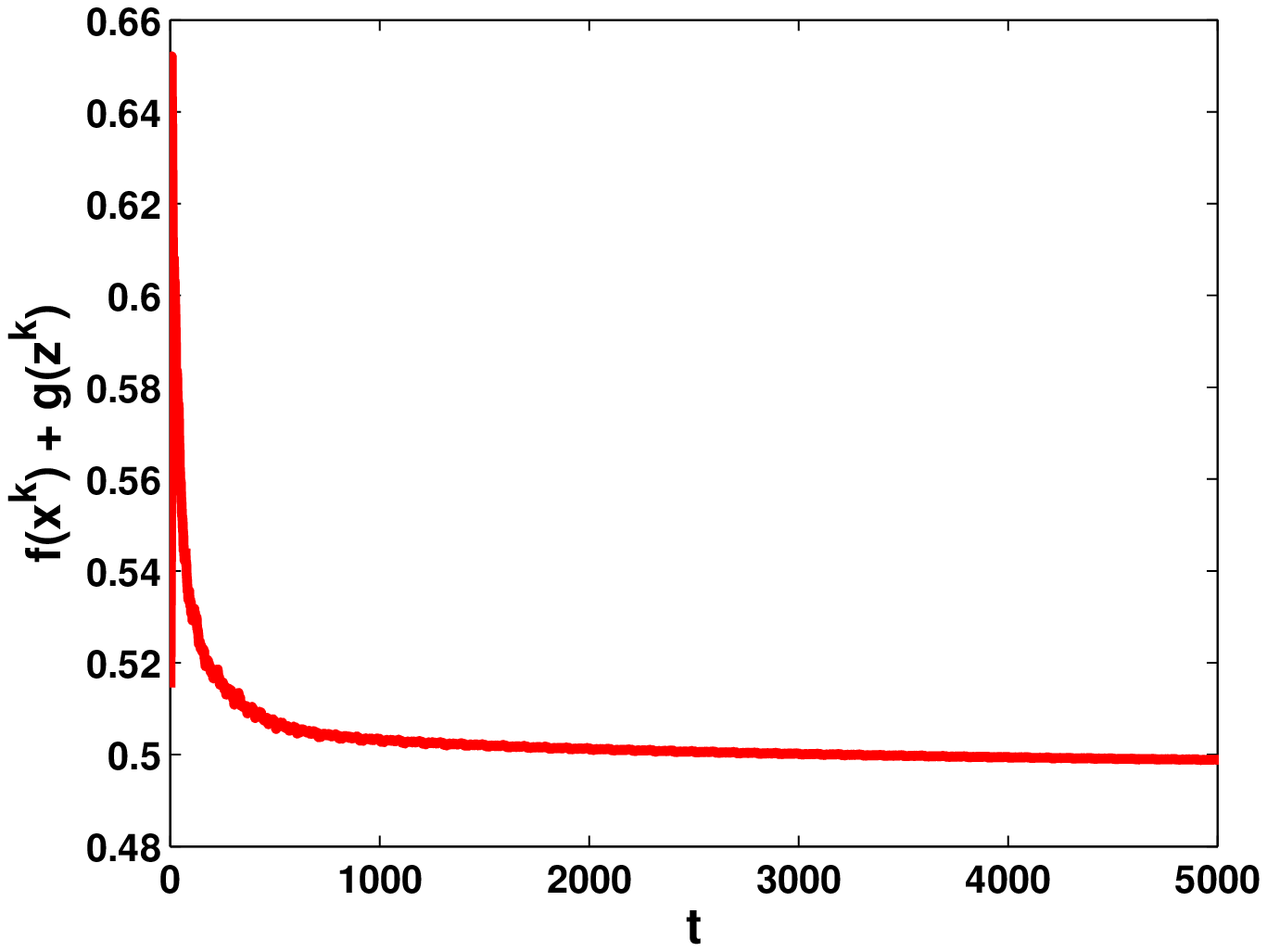}}
\subfigure[Constraints (top), primal residual (bottom). ]{\label{fig:cst_best}
\includegraphics[width = 49mm,height = 35mm]{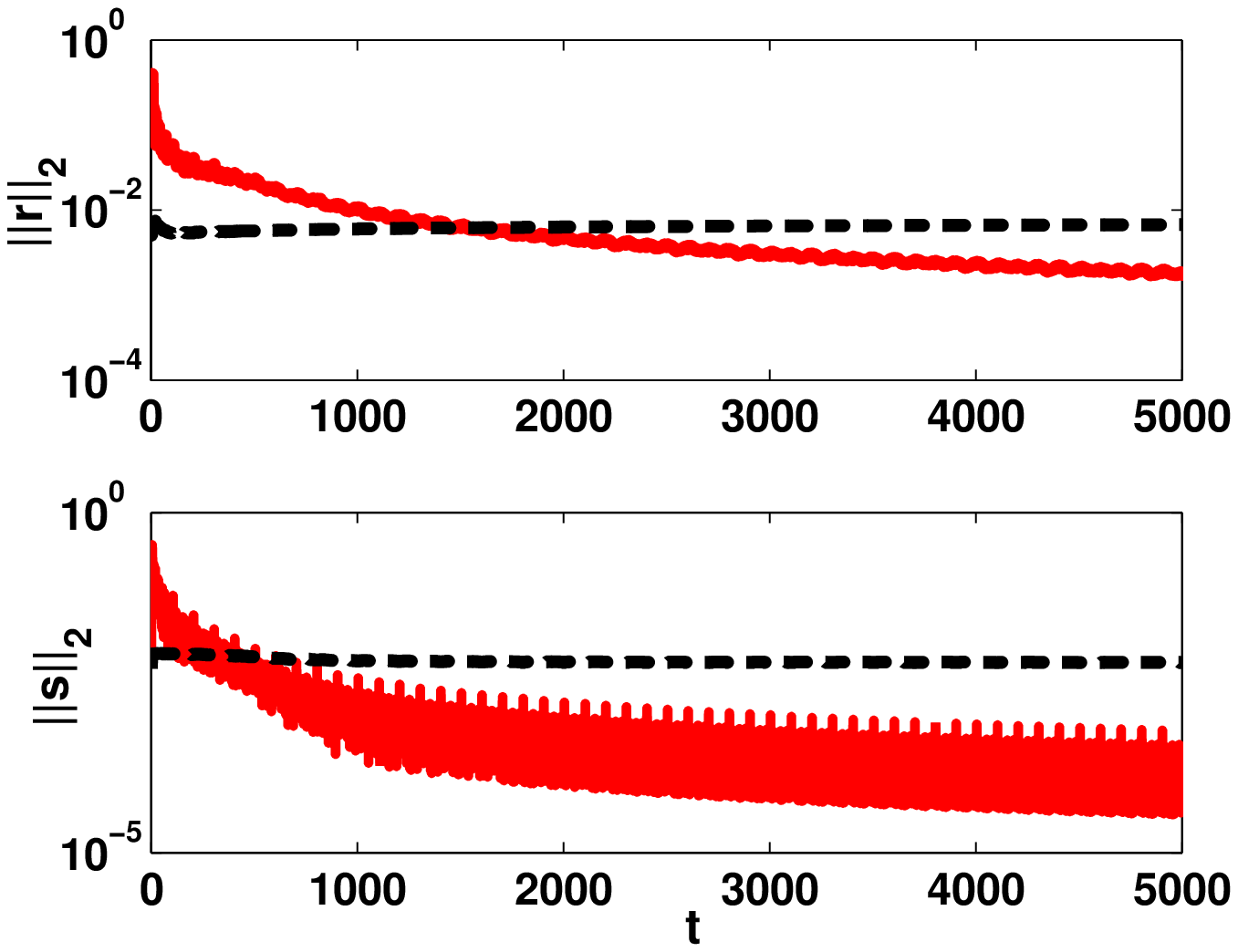}}
\caption{The convergence of sparsity, objective value and constraints for lasso in OADM with $q = 0.5,\rho = 1, \eta = t$.
} \label{fig:cvg_best}
\end{figure}

\textbf{Convergence}: We go through the examples 100 times using OADM.
Figure~\ref{fig:sparse_best} shows that NNZs converge to a value close to the actual $k = 100$ before $t = 2000$.
Figure~\ref{fig:obj_best} shows the convergence of objective value.
In Figure~\ref{fig:cst_best}, the dashed lines are the standard stopping criteria used in ADM \cite{boyd10}. Figure~\ref{fig:cst_best} shows that the equality constraint (top) and primal residual (bottom)  are satisfied in the online setting.  While the objective converges fast, the equality constraints take relatively more time to be satisfied.

\textbf{Sparsity:} We compare NNZs found by batch ADM and three online learning algorithms, including OADM, FOBOS, and RDA. We set $\eta = 1000$ for OADM and $\gamma = 1$ for RDA. For FOBOS, we use a time varying parameter $\rho_t = \rho/\sqrt{t}$. For online learning algorithms, we go through the examples 100 times. We run the experiment 20 times and the average results are plotted. We show the results for $q = 0.5$ in Figure 2, where $n$ is $1000$ for the first three figures (a)-(c) and $5000$ for the last three. While ADM and RDA tend to give the sparsest results, OADM seems more conservative and converges to reasonably sparse solutions. Figure 2 shows OADM is closest to the actual NNZs 100. The NNZs in FOBOS is large and oscillates in a big range, which has also been observed in \cite{xiao10}.

\begin{figure}[!t]\label{fig:sparsity}
\centering
\subfigure[$n = 1000, \rho = 0.1$.]{ \label{fig:sp1}
\includegraphics[width = 49mm,height = 35mm]{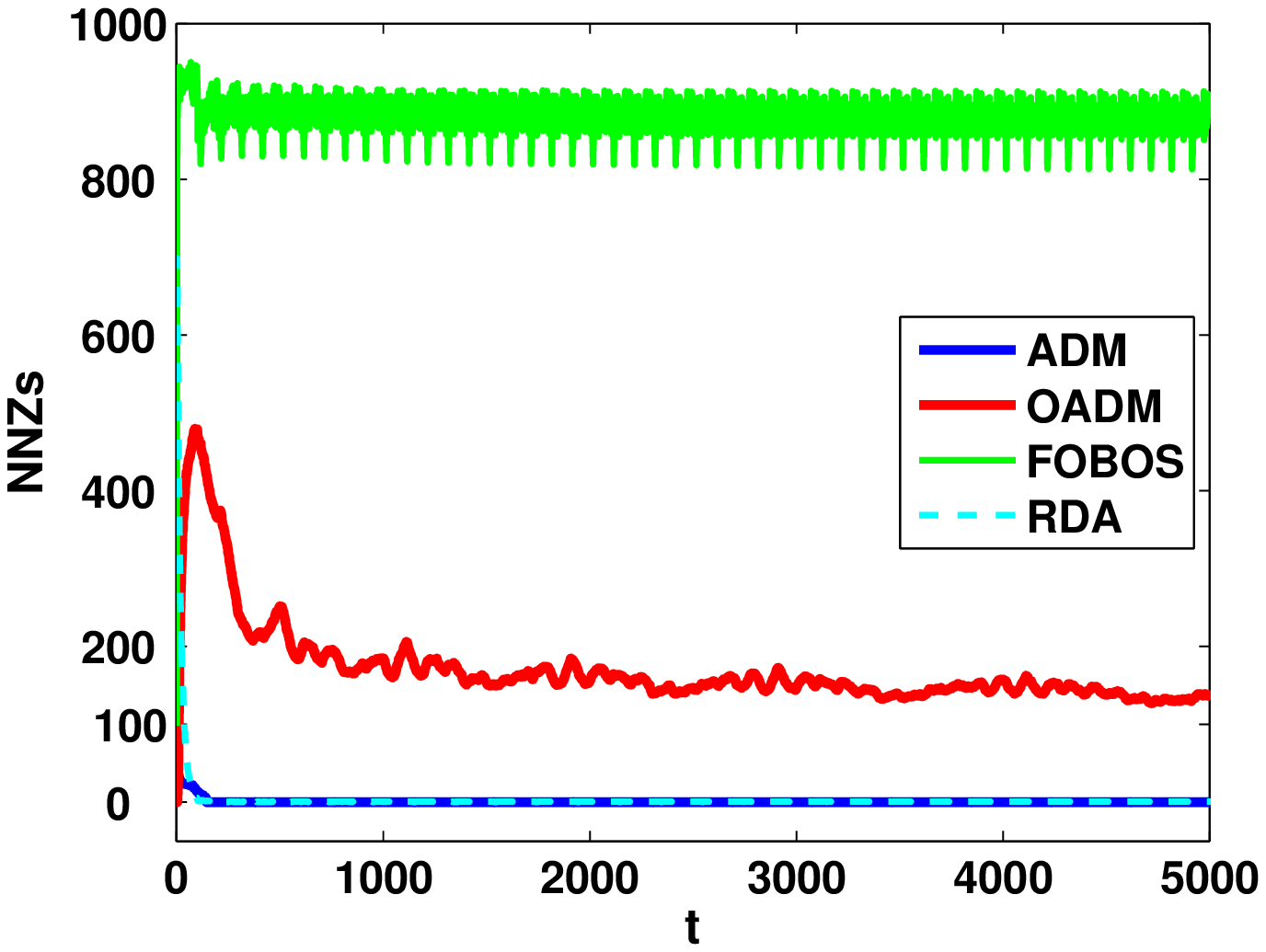}}
\subfigure[$n = 1000, \rho = 1$.]{ \label{fig:sp2}
\includegraphics[width= 49mm,height = 35mm]{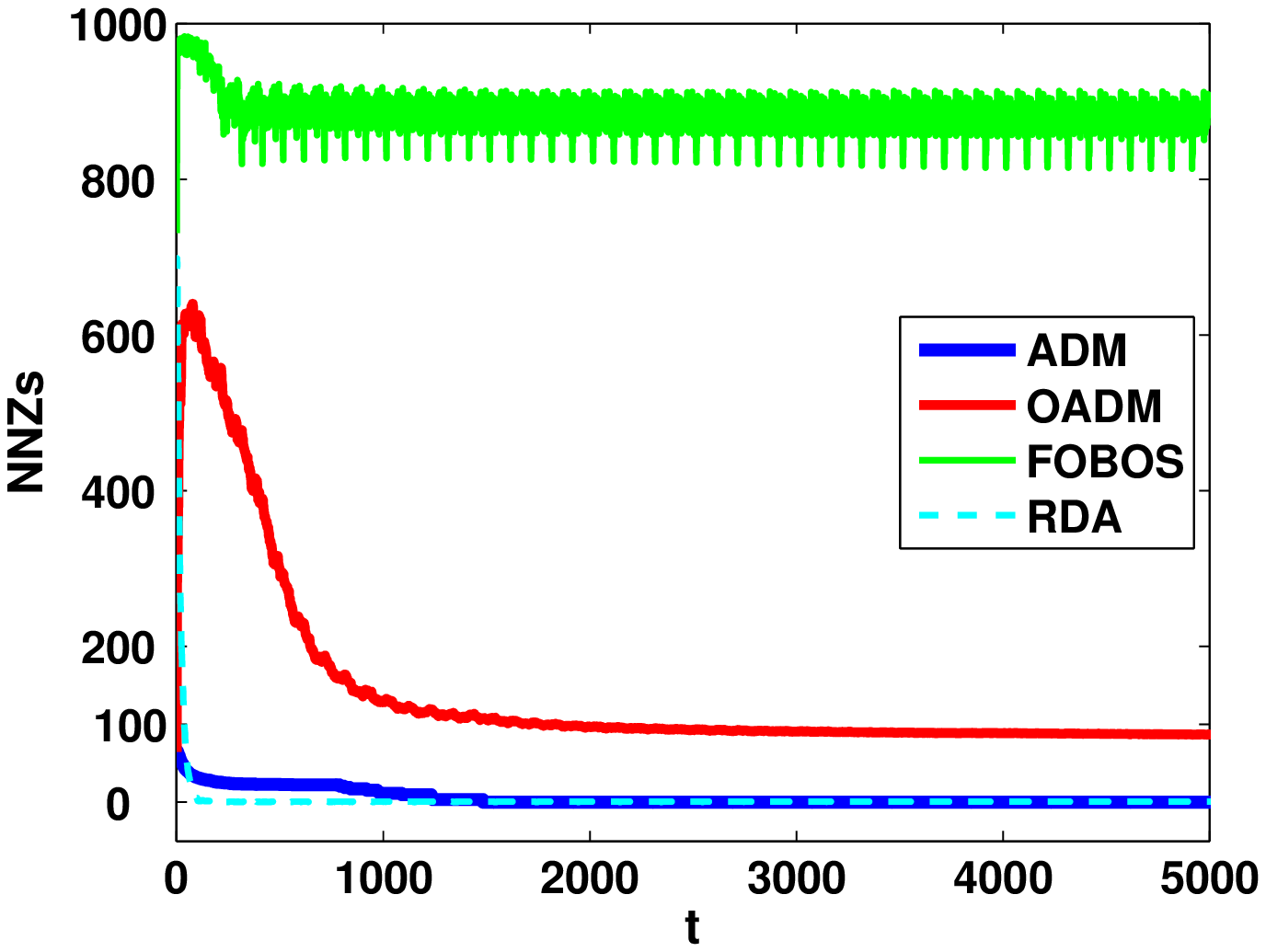}}
\subfigure[$n = 1000, \rho = 10$.]{
\label{fig:sp3}
\includegraphics[width = 49mm,height = 35mm]{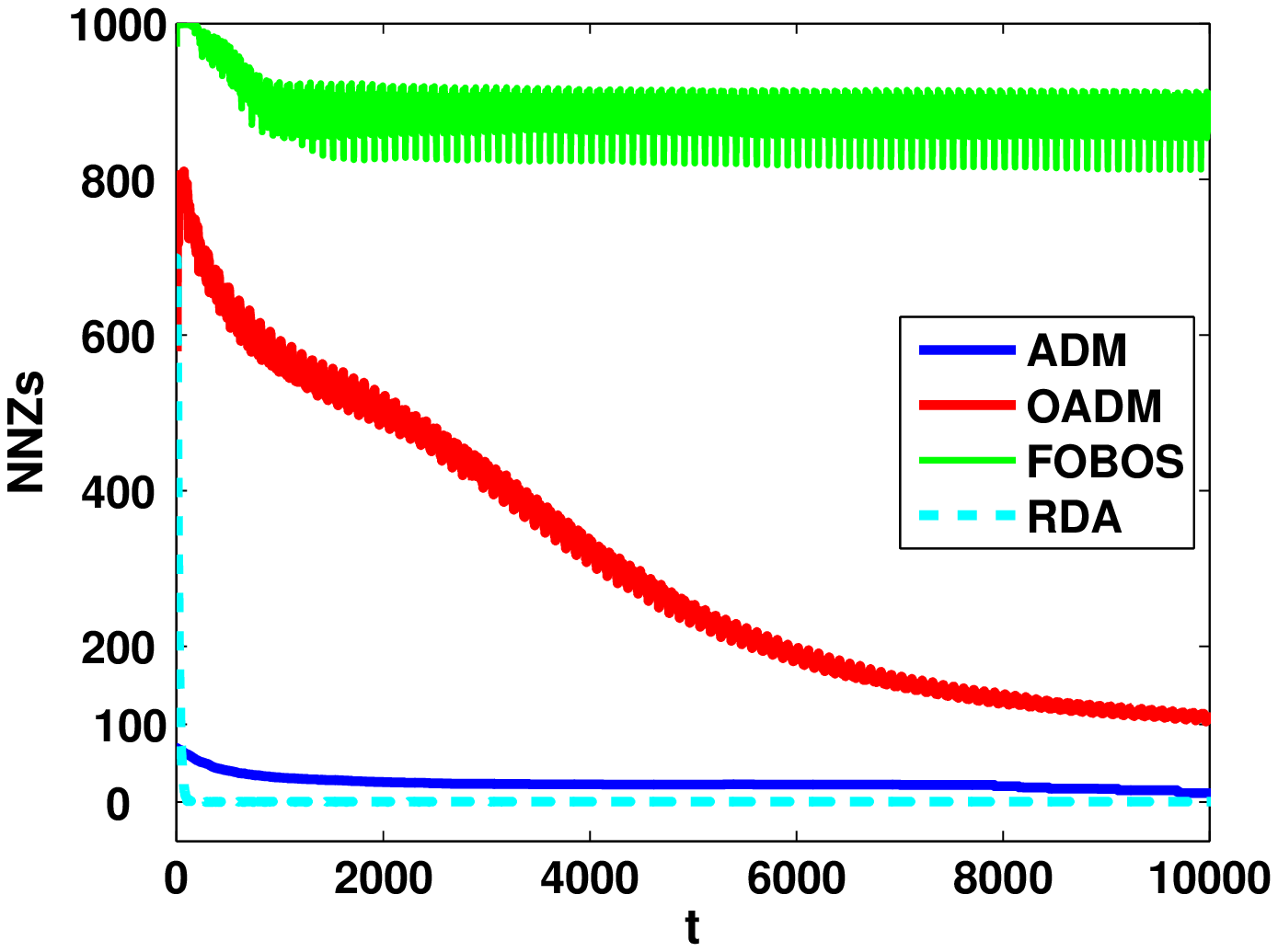}}
\subfigure[$n = 5000, \rho = 0.1$.]{ \label{fig:sp1}
\includegraphics[width = 49mm,height = 35mm]{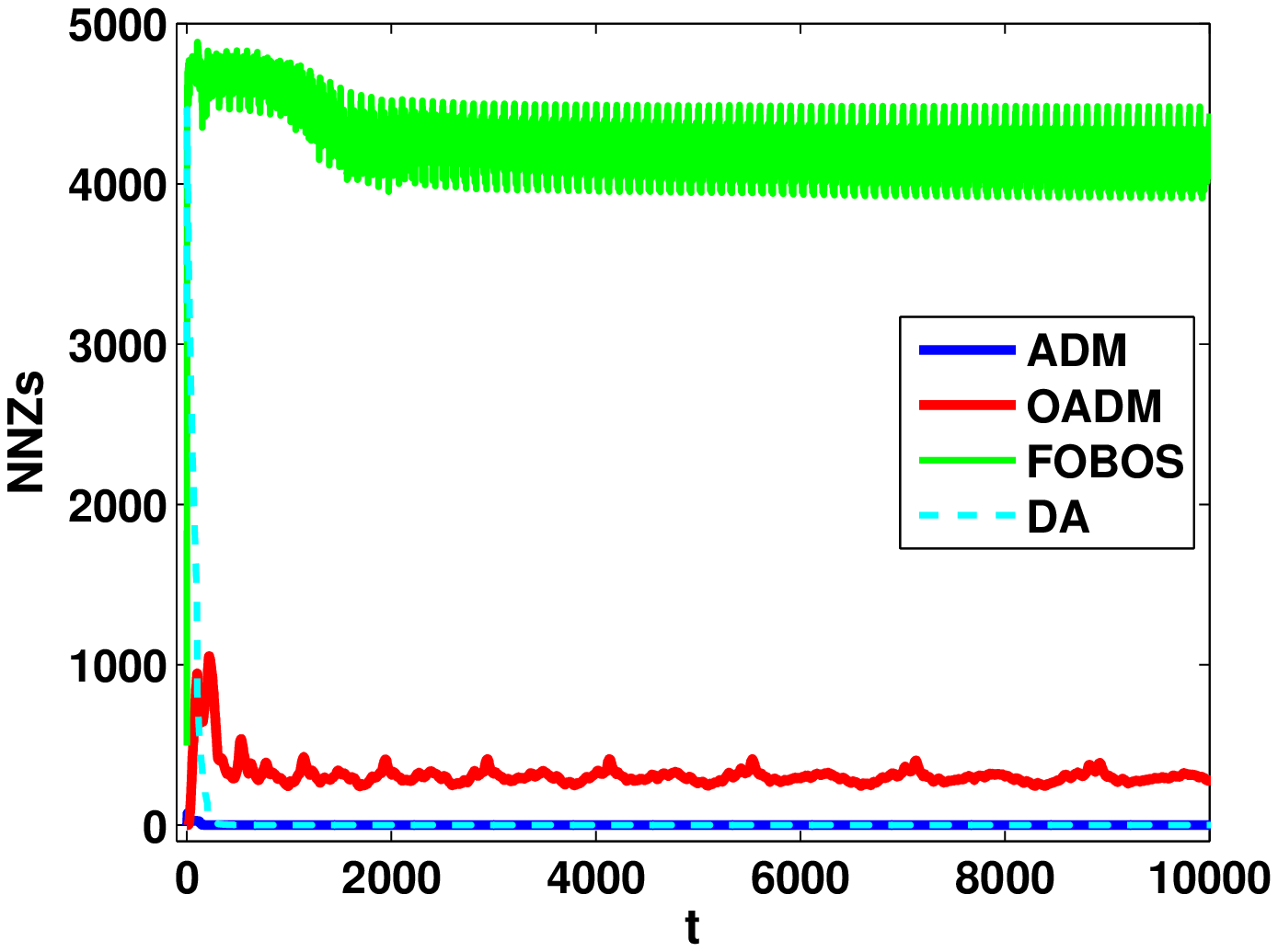}}
\subfigure[$n = 5000, \rho = 1$.]{ \label{fig:sp2}
\includegraphics[width= 49mm,height = 35mm]{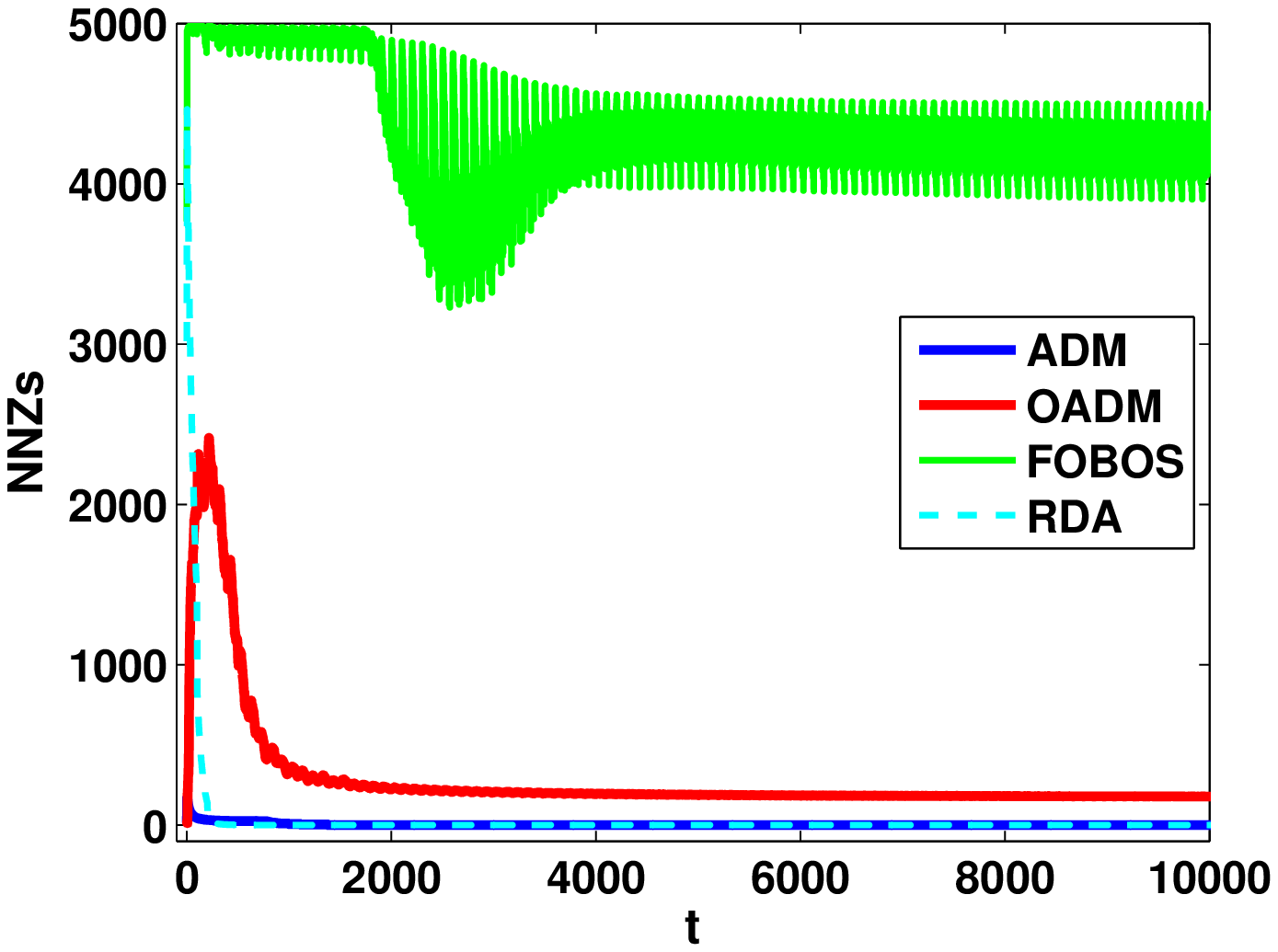}}
\subfigure[$n = 5000, \rho = 10$.]{
\label{fig:sp3}
\includegraphics[width = 49mm,height = 35mm]{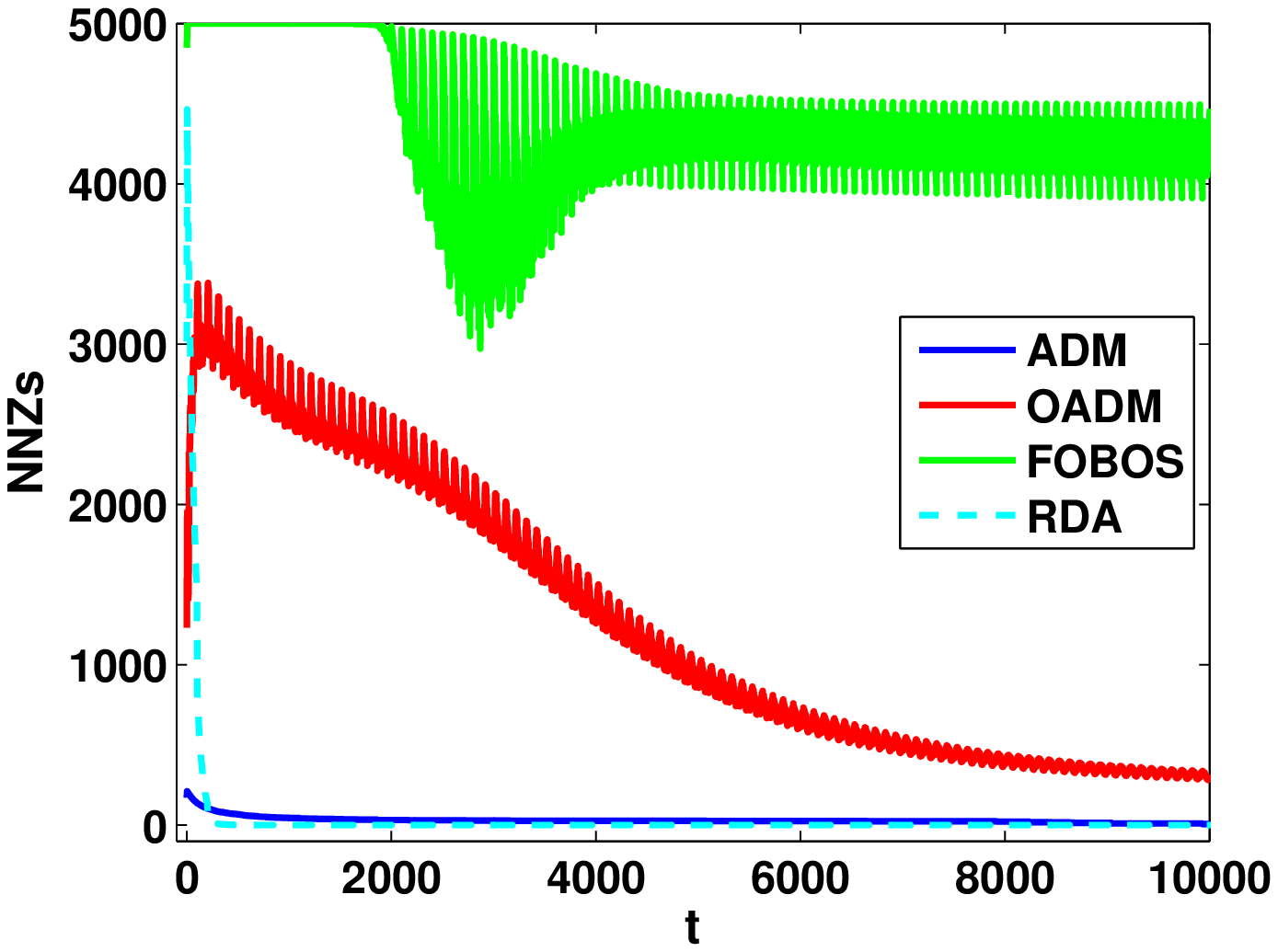}}
\caption{The NNZs found by OADM, ADM, FOBOS and RDA with $q = 0.5$ for lasso. OADM is closest to the actual NNZs.
}
\end{figure}

\begin{figure}[!th]
\centering
\subfigure{ \label{fig:tv1}
\includegraphics[width = 49mm,height = 35mm]{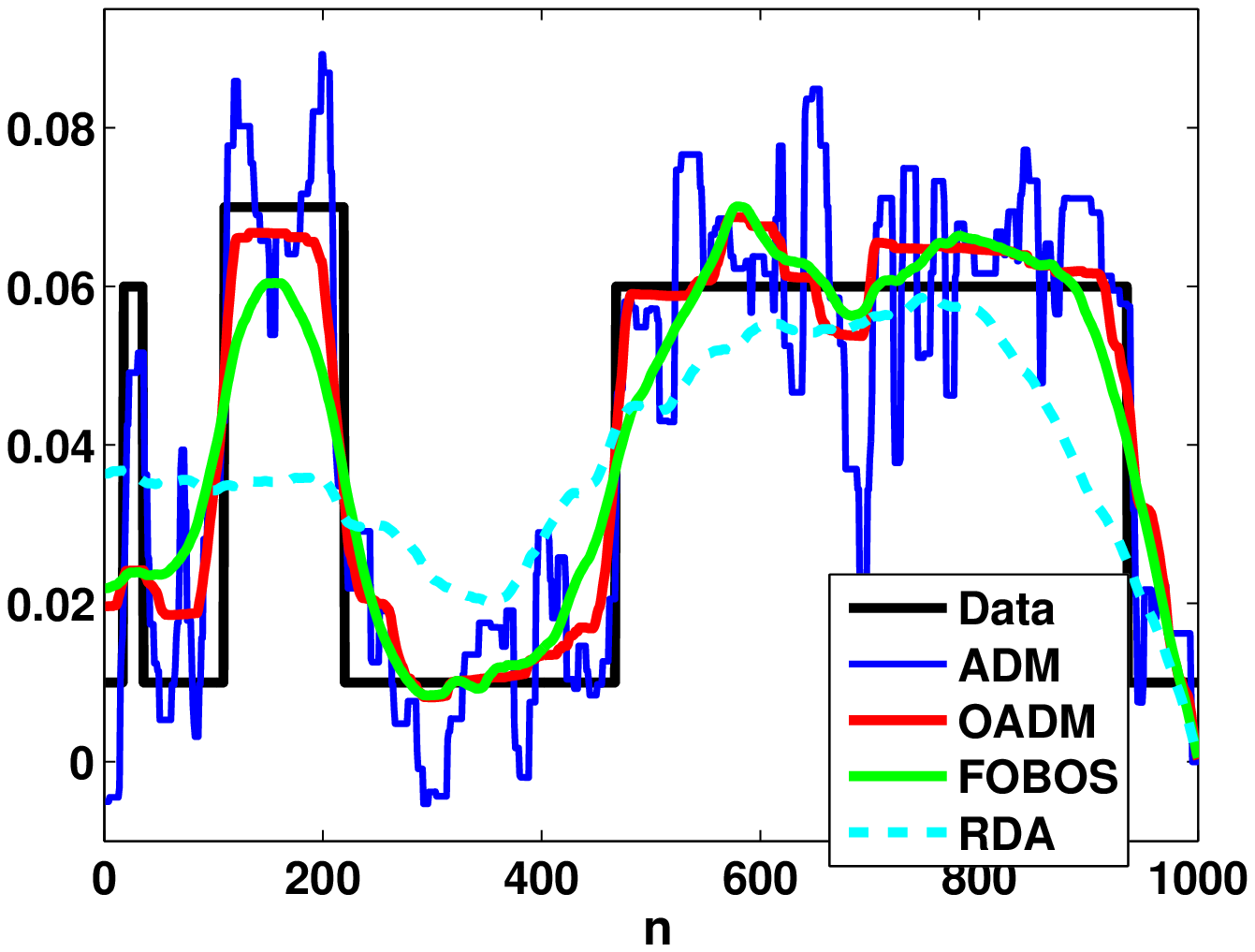}}
\subfigure{ \label{fig:tv2}
\includegraphics[width= 49mm,height = 35mm]{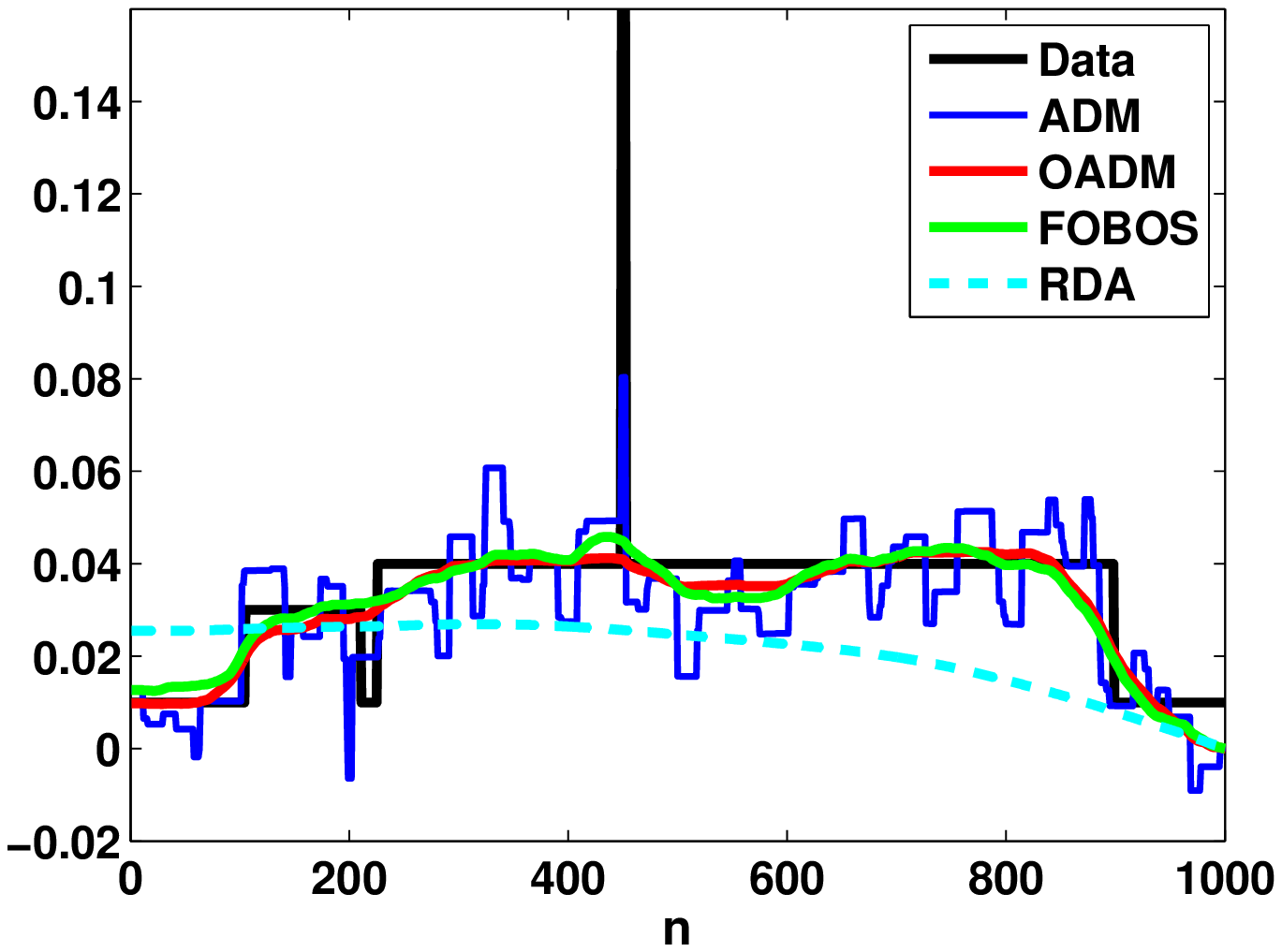}}
\subfigure{
\label{fig:tv3}
\includegraphics[width = 49mm,height = 35mm]{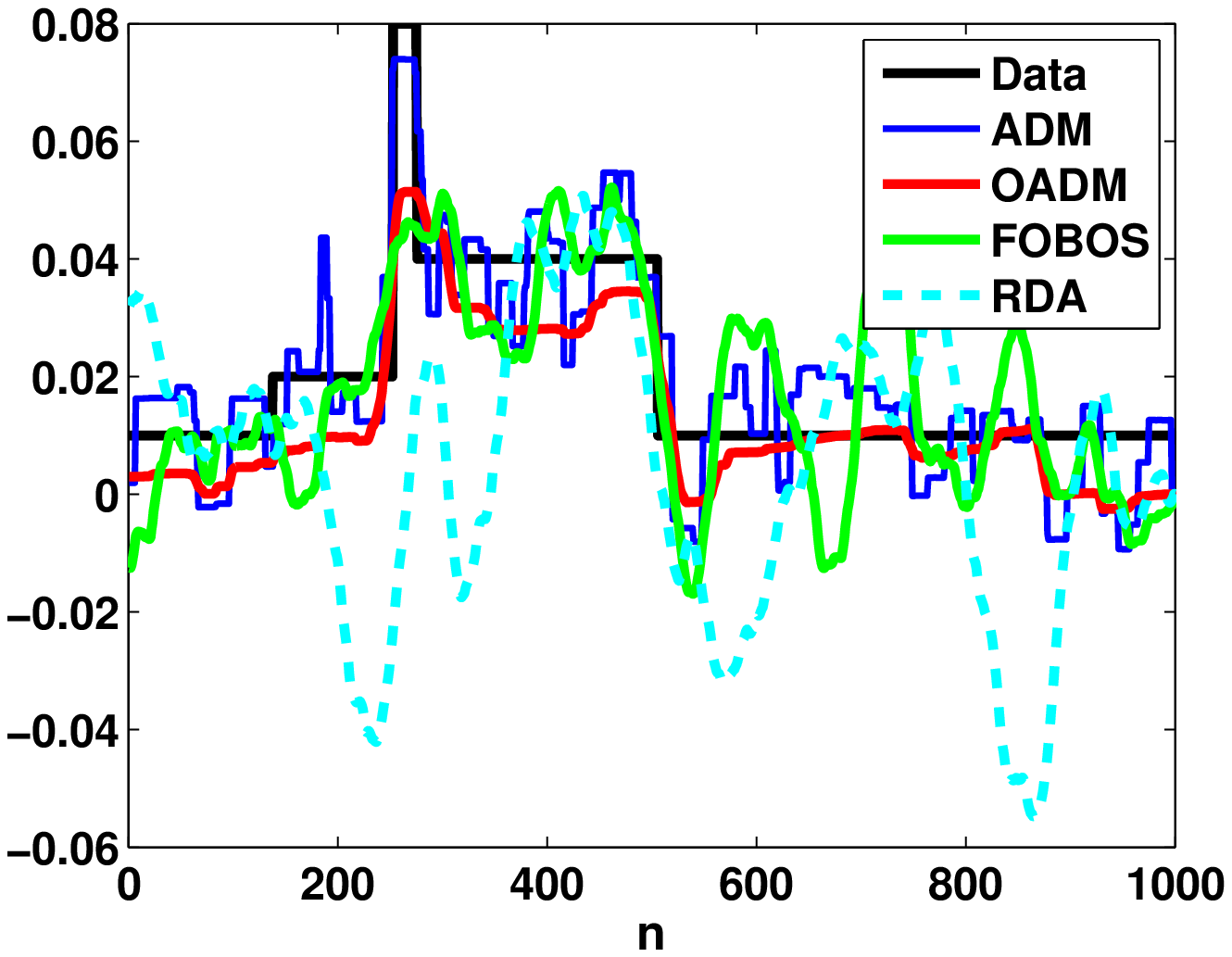}}
\caption{The TV patterns found by OADM, ADM, FOBOS and RDA. OADM is the best in recovering the patterns.
}
\label{fig:tv}
\end{figure} 

\textbf{Total Variation:} We compare the patterns found by the four algorithms. For all algorithms, $N= 100, n = 1000, \lambda = 0.001$ and $\rho$ is chosen through cross validation. In RDA, $\gamma = 100$. Recall that $\eta = 0$ in OADM. While we use a fixed $\rho$ for OADM and RDA, FOBOS uses $\rho_t = \rho/\sqrt{t}$. Figure~\ref{fig:tv} shows the three different patterns and results found by the algorithms.
ADM seems to follow the pattern with oscillation. OADM is smoother and generally follows the trend of the patterns. For the first two examples, FOBOS works well and the patterns found by RDA tend to be flat. In the last example, both FOBOS and RDA oscillate.

\section{Conclusions}
In this paper, we first developed new proof techniques to analyze the convergence rate for ADM, which establishes a $O(1/T)$ convergence rate for the objective, the optimality conditions (constraints) and the variational inequality form of ADM. The new proof techniques may facilitate the improvement and modifications of ADM which is needed in some scenarios. For example, the quadratic penalty term in the $\x$ and $\z$ updates may not lead to efficient algorithm, while other Bregman divergences like KL divergence may induce efficient updates. 

We propose an efficient online learning algorithm named online ADM (OADM). Using the proof technique developed for batch ADM, we establish regret bounds for the objective and constraint violation for general and strongly convex functions in OADM. We also discuss inexact update to yield efficient $\x$ update, including mirror descent and composite objective mirror descent. Finally, we illustrate the efficacy of OADM in solving lasso and total variation problems. Through splitting variables, we show OADM can do projection-free online learning with linear constraints. It would be interesting to explore whether OADM can do projection-free learning with other constraints. Through variables splitting, ADM has been successfully used in distributed optimization. If distributed ADM is extended to the online learning setting, distributed OADM will allow the data to be distributed along the time dimension, which can be particularly useful for spatio-temporal data. 

\section*{Acknowledgment}

The research was supported by NSF CAREER award IIS-0953274, and NSF grants IIS-0916750, IIS-0812183, and IIS-1029711. The authors thank the detailed and insightful comments from reviewers and extend out thanks to Daniel Boley and Stephen Wright for helpful discussion. The authors also thank Bingsheng He and Xiaoming Yuan for pointing out the relationship with their proof techniques.

\bibliographystyle{plain}
\bibliography{all,admm,onlinelearn}

\appendix
\section{Proof of Theorem~\ref{thm:oadmm1_rgtbd} and~\ref{thm:oadm_logrgtbd1} in Case 2 in Section~\ref{sec:inexactADMM}}\label{sec:app1}
\textbf{Proof of Theorem~\ref{thm:oadmm1_rgtbd}} The first order derivative is 0, i.e.,
\begin{align}
f_t'(\x_{t}) + \bA^T \{\y_t + \rho\bA^T(\bA\x_{t} - \bB\z_t - \c)\}
+ \eta(\nabla \phi(\x_{t+1}) - \nabla \phi(\x_t)) = 0~,
\end{align}
Rearranging the terms yields
\begin{align}\label{eq:ftsg}
 -\bA^T (\y_t + \rho\bA^T(\bA\x_{t+1} - \bB\z_t - \c))
- \eta(\nabla \phi(\x_{t+1}) - \nabla \phi(\x_t))  = f_t'(\x_{t})~,
\end{align}
where the left hand side is same as~\myref{eq:oadmm1_fg}. Therefore,  $\langle f_t'(\x_t), \x_{t+1} - \x^*\rangle +  g(\z_{t+1}) - g(\z^*)$ can be written as the right hand side of~\myref{eq:oadmm1_bd1}. Using the convexity of $f_t$, we have
\begin{align}\label{eq:loadm2_ft}
f_t(\x_{t}) + &g(\z_{t+1}) - (f_t(\x^*)+g(\z^*)) \leq \langle f_t'(\x_t), \x_t - \x^* \rangle +  g(\z_{t+1}) - g(\z^*) \nonumber \\
&= \langle f_t'(\x_t), \x_{t+1} - \x^*\rangle +  g(\z_{t+1}) - g(\z^*) + \langle f_t'(\x_t), \x_t - \x_{t+1} \rangle~.
\end{align}
Applying~\myref{eq:oadmm1_fb0} for the last term, we have~\myref{eq:oadm1_r0}. Therefore, Theorem~\ref{thm:oadmm1_rgtbd} holds for Case 2. 

\textbf{Proof of Theorem~\ref{thm:oadm_logrgtbd1}}
Using the strong convexity of $f_t$ and $g$ defined in~\myref{eq:fsc} and~\myref{eq:gsc} respectively, we have
\begin{align}
&f_t(\x_{t}) + g(\z_{t+1})- (f_t(\x^*) + g(\z^*)) \nonumber \\
&\leq \langle f_t'(\x_t), \x_{t} - \x^*\rangle - \beta_1 B_{\phi}(\x^*,\x_t)+ \langle g'(\z_{t+1}), \z_{t+1} - \z^*\rangle - \frac{\beta_2}{2} \| \z^* - \z_{t+1} \|_2^2 \nonumber \\
&=   \langle f_t'(\x_t), \x_{t+1}-\x^*\rangle + \langle f_t'(\x_t), \x_t-\x_{t+1} \rangle+ \langle g'(\z_{t+1}), \z_{t+1}-\z^*\rangle  - \beta_1 B_{\phi}(\x^*,\x_t) - \frac{\beta_2}{2} \| \z^*-\z_{t+1} \|_2^2~.
\end{align}
The first four terms are the same as in~\myref{eq:loadm2_ft}, which can be reduced to~\myref{eq:oadm1_r0}.  Therefore, adding the last two terms to~\myref{eq:oadm1_r0}, we have
\begin{align}
&f_t(\x_{t}) + g(\z_{t+1}) - (f_t(\x^*) + g(\z^*)) \nonumber \\
& \leq \frac{1}{2\rho}(\| \y_{t} \|_2^2 - \| \y_{t+1} \|_2^2)
- \frac{\rho}{2} \|\bA\x_{t+1} + \bB\z_t - \c\|_2^2
+ \frac{\rho}{2}  ( \|\bB\z^* - \bB\z_t\|_2^2 -  \|\bB\z^* - \bB\z_{t+1}\|_2^2) \nonumber \\
& \quad + \frac{1}{2\alpha\eta} \| f_t'(\x_{t}) \|_q^2  + \eta (B_{\phi}(\x^*,\x_t) - B_{\phi}(\x^*, \x_{t+1}))  - \beta_1 B_{\phi}(\x^*,\x_t) - \frac{\beta_2}{2} \| \z^*-\z_{t+1} \|_2^2~.
\end{align}
Summing over $t$ from $1$ to $T$, we have
\begin{align}\label{eq:oiadm_r1}
R_1(T) & \leq \sum_{t=1}^T  \frac{1}{2\rho_{t+1}}(\| \y_{t} \|_2^2 - \| \y_{t+1} \|_2^2) + \frac{1}{2\beta} \sum_{t=0}^T  \frac{1}{\eta_{t+1}}\| f_t'(\x_{t}) \|_2^2 \nonumber \\
& + \sum_{t=1}^T (\frac{\rho_{t+1}}{2}  ( \|\bB\z^* - \bB\z_t\|_2^2 -  \|\bB\z^* - \bB\z_{t+1}\|_2^2) - \frac{\beta_2}{2}\|\z^* - \z_{t+1}\|_2^2) \nonumber \\
& + \sum_{t=1}^T (\eta_{t+1} (B_{\phi}(\x^*,\x_t) - B_{\phi}(\x^*, \x_{t+1})) - \beta_1 B_{\phi}(\x^*, \x_{t}) )~.
\end{align}
The difference between~\myref{eq:oiadm_r1} and~\myref{eq:R1_logrgt} lies in the last term. Setting $\eta_t = \beta_1 t$, we have the following telescoping sum for the last term :
\begin{align}
&\sum_{t=1}^T (\eta_{t+1} (B_{\phi}(\x^*,\x_t) - B_{\phi}(\x^*, \x_{t+1})) - \beta_1 B_{\phi}(\x^*, \x_{t}) ) \nonumber \\
& \leq \eta_2B_{\phi}(\x^*,\x_1) + \sum_{t=2}^T B_{\phi}(\x^*,\x_t)(\eta_{t+1} - \eta_{t} -\beta_1 ) \nonumber \\
& = 2\beta_1 D^2_\x~,
\end{align}
which is the same as~\myref{eq:logrgt_phi}.  Therefore, Theorem~\ref{thm:oadm_logrgtbd1} holds for the Case 2.

\section{Proof of Stochastic Convergence Rates}\label{sec:app2}
Although the proof is based on Case 2 in Section 6.1, Case 3 and 4 will follow automatically. In the stochastic setting, replacing $f_t'(\x_{t})$ by $f'(\x_{t},\xi_t)$ in~\myref{eq:ftsg} gives
\begin{align}
 -\bA^T (\y_t + \rho\bA^T(\bA\x_{t+1} - \bB\z_t - \c))
- \eta(\nabla \phi(\x_{t+1}) - \nabla \phi(\x_t))  = f'(\x_{t},\xi_t)~,
\end{align}
(a) Replacing $f_t(\x_{t}), f'_t(\x_t)$ by $f(\x_{t}),f'(\x_t,\xi_t)$ respectively in~\myref{eq:loadm2_ft} gives 
\begin{align}\label{eq:sadmm_fg001}
& f(\x_{t}) + g(\z_{t+1}) - (f(\x^*)+g(\z^*)) \leq \langle f'(\x_t,\xi_t), \x_t - \x^* \rangle +  g(\z_{t+1}) - g(\z^*) \nonumber \\
&= \langle f'(\x_t,\xi_t), \x_{t+1} - \x^*\rangle +  g(\z_{t+1}) - g(\z^*) + \langle f'(\x_t,\xi_t), \x_t - \x_{t+1} \rangle~.
\end{align}
As a result, we have the following result by replacing $f_t(\x_{t}), f'_t(\x_t)$ by $f(\x_{t}),f'(\x_t,\xi_t)$ in~\myref{eq:oadm1_r0}
\begin{align}
&f(\x_{t}) + g(\z_{t+1}) - (f(\x^*) + g(\z^*)) \nonumber \\
& \leq \frac{1}{2\rho}(\| \y_{t} \|_2^2 - \| \y_{t+1} \|_2^2)
+ \frac{\rho}{2}  ( \|\bB\z^* - \bB\z_t\|_2^2 -  \|\bB\z^* - \bB\z_{t+1}\|_2^2) - \frac{\rho}{2} \|\bA\x_{t+1} + \bB\z_t - \c\|_2^2 \nonumber \\
& \quad + \frac{1}{2\alpha\eta} \| f'(\x_{t},\xi_t) \|_q^2  + \eta (B_{\phi}(\x^*,\x_t) - B_{\phi}(\x^*, \x_{t+1})) ~.
\end{align}
Moving the term $\frac{\rho}{2} \|\bA\x_{t+1} + \bB\z_t - \c\|_2^2$  to the left hand side and using Lemma \ref{lem:pdr},  we have
\begin{align}
&f(\x_{t}) + g(\z_{t+1}) - (f_t(\x^*) + g(\z^*)) + \frac{\rho}{2} \|\bA\x_{t+1} + \bB\z_{t+1} - \c\|_2^2 \nonumber \\
& \leq \frac{1}{2\rho}(\| \y_{t} \|_2^2 - \| \y_{t+1} \|_2^2)
+ \frac{\rho}{2}  ( \|\bB\z^* - \bB\z_t\|_2^2 -  \|\bB\z^* - \bB\z_{t+1}\|_2^2) \nonumber \\
& \quad + \frac{1}{2\alpha\eta} \| f'(\x_{t},\xi_t) \|_q^2  + \eta (B_{\phi}(\x^*,\x_t) - B_{\phi}(\x^*, \x_{t+1})) ~.
\end{align}
 Summing over $t$ from $0$ to $T-1$ and following the derivation in~\myref{eq:oadm1_r01}, we have
 \begin{align}
&\sum_{t=1}^T \left[ f(\x_{t}) + g(\z_{t}) - (f(\x^*) + g(\z^*)) + \frac{\rho}{2} \|\bA\x_{t+1} + \bB\z_{t+1} - \c\|_2^2 \right] \leq \frac{\lambda_{\max}^{\bB}D_{\z}^2\rho}{2} + \eta D_{\x}^2 + \frac{\| f'(\x_{t},\xi_t) \|_q^2T}{2\alpha \eta}~.
 \end{align}
  Dividing both sides by $T$, applying the Jensen's inequality, we have
    \begin{align}\label{eq:sadmm_objcst0}
f(\bar{\x}_T)+g(\bar{\z}_T) - (f(\x^*) + g(\z^*)) + \frac{\rho}{2}\| \bA\bar{\x}_{T} + \bB\bar{\z}_{T} + \c\|_2^2 \leq \frac{\lambda_{\max}^{\bB}D_{\z}^2\rho}{2T}  + \frac{\eta D_{\x}^2}{T} + \frac{\| f'(\x_{t},\xi_t) \|_q^2}{2\alpha \eta}~.
     \end{align}  
Assume $\E [\| f'(\x_{t},\xi_t) \|_q^2] \leq G_f^2$.  Setting $\eta = \frac{G_f\sqrt{T}}{D_\x \sqrt{2\alpha}}$ and taking expectation, we have
  \begin{align}\label{eq:sadmm_objcst}
   \E\left[f(\bar{\x}_T)+g(\bar{\z}_T) - (f(\x^*) + g(\z^*)) + \frac{\rho}{2}\| \bA\bar{\x}_{T} + \bB\bar{\z}_{T} + \c\|_2^2\right] \leq \frac{\lambda_{\max}^{\bB}D_{\z}^2\rho}{2T} + \frac{\sqrt{2}G_fD_\x}{\sqrt{\alpha}\sqrt{T}}~.
   \end{align}  
~\myref{eq:sadmm_eobj} follows by setting $\rho = \sqrt{T}$.

Assume $f(\bar{\x}_T)+g(\bar{\z}_T) - (f(\x^*) + g(\z^*)) \geq -F$. Dividing both sides by $\frac{\rho}{2}$ and rearranging the terms yield
  \begin{align}\label{eq:sadmm_cst0}
   \E\left[\| \bA\bar{\x}_{T} + \bB\bar{\z}_{T} + \c\|_2^2\right] \leq \frac{2F}{\rho} + \frac{\lambda_{\max}^{\bB}D_{\z}^2}{T} + \frac{2\sqrt{2}G_fD_\x}{\rho \sqrt{\alpha}\sqrt{T}}~.
   \end{align}
Setting $\rho = \sqrt{T}$ gives~\myref{eq:sadmm_ecst}.

(b) Using the convexity of $f$, we have
\begin{align}\label{eq:sadmmf}
f(\x_t) - f(\x^*) \leq \langle f'(\x_t), \x_t - \x^*\rangle = \langle f'(\x_t,\xi_t),\x_{t+1} - \x^*\rangle + \langle f'(\x_t,\xi_t),\x_t-\x_{t+1}\rangle + \epsilon_t~.
\end{align}
where 
\begin{align}
\epsilon_t  = \langle f'(\x_t) - f'(\x_t,\xi_t),\x_t - \x^*\rangle~.
\end{align}
Let $\mathcal{F}$ be a filtration with $\xi_t \in\mathcal{F}_t$ for $t \leq T$. Since $\x_t\in\mathcal{F}_{t-1}$,
\begin{align}
\E[\epsilon_t|\mathcal{F}_{t-1}] = \langle f'(\x_t) - \E[f'(\x_t,\xi_t)|\mathcal{F}_{t-1}],\x_t - \x^*\rangle = 0~.
\end{align}
Therefore, $\sum_{t=1}^T \epsilon_t$ is a martingale difference sequence.  Assuming $B_{\phi}(\x^*,\x^t) \leq D_{\x}^2$,  $\|\x_t - \x^*\|_p \leq \sqrt{\frac{2}{\alpha}} D_\x$. We have
\begin{align}
|\epsilon_t| \leq \|f'(\x_t) - f'(\x_t,\xi_t)\|_q \|\x_t - \x^*\|_p \leq 2\sqrt{\frac{2}{\alpha}} D_\x G_{f}~.
\end{align}
Applying Azuma-Hoeffding inequality~\cite{azuma} on $\sum_{t=1}^T \epsilon_t$ yields
\begin{align}\label{eq:azuma}
P(\sum_{t=1}^T \epsilon_t \geq \varepsilon) \leq \exp\left(-\frac{\alpha\varepsilon^2}{16TD^2_\x G_f^2}\right)~.
\end{align}
Combing~\myref{eq:sadmm_fg001} and~\myref{eq:sadmmf}, we have
 \begin{align}\label{eq:sadmm_cvg00}
 f(\x_{t}) + &g(\z_{t+1}) - (f(\x^*)+g(\z^*)) \leq \langle f_t'(\x_t), \x_t - \x^* \rangle +  g(\z_{t+1}) - g(\z^*) \nonumber \\
 &=  \langle f'(\x_t,\xi_t),\x_{t+1} - \x^*\rangle +  g(\z_{t+1}) - g(\z^*)  + \langle f'(\x_t,\xi_t),\x_t-\x_{t+1}\rangle + \epsilon_t~.
  \end{align}
As a result, \myref{eq:sadmm_objcst0} becomes
      \begin{align}
  & f(\bar{\x}_T)+g(\bar{\z}_T) - (f(\x^*) + g(\z^*)) + \frac{\rho}{2}\| \bA\bar{\x}_{T} + \bB\bar{\z}_{T} + \c\|_2^2 \nonumber \\
  &\leq \frac{\lambda_{\max}^{\bB}D_{\z}^2\rho}{2T}  + \frac{\eta D_{\x}^2}{T} + \frac{\| f'(\x_{t},\xi_t) \|_q^2}{2\alpha \eta}+ \frac{1}{T}\sum_{t=1}^T \epsilon_t~.
       \end{align}  
Assuming $\| f'(\x_{t},\xi_t) \|_q \leq G_f$ and setting $\eta = \frac{G_f\sqrt{T}}{D_\x \sqrt{2\alpha}}, \rho = \sqrt{T}$, we have
 \begin{align}\label{eq:sadmm_objcsth}
&f(\bar{\x}_{T}) + g(\bar{\z}_{T}) - (f(\x^*) + g(\z^*)) + \frac{\rho}{2}\| \bA\bar{\x}_{T} + \bB\bar{\z}_{T} + \c\|_2^2 \leq \frac{\lambda_{\max}^{\bB}D_{\z}^2}{2\sqrt{T}} + \frac{\sqrt{2}G_fD_\x}{\sqrt{\alpha}\sqrt{T}} + \frac{1}{T}\sum_{t=1}^T \epsilon_t~.
 \end{align}
Applying~\myref{eq:azuma} gives~\myref{eq:sadm2_bd}.

Assume $f(\bar{\x}_T)+g(\bar{\z}_T) - (f(\x^*) + g(\z^*)) \geq -F$. In~\myref{eq:sadmm_objcsth}, dividing both sides by $\frac{\rho}{2} = \frac{\sqrt{T}}{2}$ and rearranging the terms yield
  \begin{align}\label{eq:sadmm_cst0}
 \| \bA\bar{\x}_{T} + \bB\bar{\z}_{T} + \c\|_2^2 \leq \frac{2F}{\rho} + \frac{\lambda_{\max}^{\bB}D_{\z}^2}{T} + \frac{2\sqrt{2}G_fD_\x}{ \sqrt{\alpha}T}+ \frac{1}{T}\sum_{t=1}^T\epsilon_t~.
   \end{align}
Applying~\myref{eq:azuma} yields~\myref{eq:sadm2_cst}.
\qed

\end{document}